%% file: manuscript.tex
\documentclass[11pt,a4paper]{article}
\usepackage[margin=23mm]{geometry}
\usepackage[T1]{fontenc}
\usepackage{lmodern,microtype,amsmath,amssymb,graphicx,booktabs,array,longtable}
\usepackage{xcolor,fancyhdr,enumitem,placeins,caption,float}
\usepackage[colorlinks=true,linkcolor=blue!45!black,citecolor=blue!45!black,urlcolor=blue!45!black]{hyperref}
\hypersetup{pdftitle={Humanoid Locomotion with a Fly-Inspired Recurrent Controller},pdfauthor={Isabel Guan, Yuntian Zhao, Dingyuan Zhang, Shipeng Lyu}}
\setlist[itemize]{noitemsep,topsep=3pt,leftmargin=18pt}
\input{results.tex}
\newcommand{\shot}[1]{\includegraphics[width=.475\linewidth]{screenshots/#1.png}}
\title{\vspace{-1.2cm}\LARGE\bfseries Humanoid Locomotion with a\\Fly-Inspired Recurrent Controller}
\author{\normalsize Isabel Guan$^{1}$, Yuntian Zhao$^{2}$, Dingyuan Zhang$^{3}$, Shipeng Lyu$^{4}$\\[5pt]
\small $^{1}$The Hong Kong University of Science and Technology, Hong Kong\\
\small $^{2}$Zenbot\\
\small $^{3}$Nanyang Technological University, Singapore\\
\small $^{4}$The Hong Kong Polytechnic University, Hong Kong}
\date{23 September 2026}
\begin{document}
\maketitle
\vspace{-7mm}
\begin{center}
{\small\color{black!65}Preprint\quad|\quad Simulation study\quad|\quad Not peer reviewed}
\end{center}
\begin{abstract}
We investigate humanoid locomotion with a fly-inspired recurrent controller and identify the pathways supporting its deployed behavior. The controller couples 3,609 continuous neural states to a simulated Unitree G1 through body-observation projections, a motor-neuron-labelled readout, and joint servos. We formulate this neural--body feedback system and evaluate a fixed checkpoint across seven terrain instances, three speeds, and three initial yaw offsets. It completes \GraphSuccess/63 conditions under a survival-and-forward-progress criterion; a privileged reference completes \TeacherSuccess/63. At nominal yaw, resetting the recurrent motor state before every policy call changes success from \GraphNominalSuccess/21 to \ResetSuccess/21. Conversely, depth and upstream-state substitutions at \ProbeCount\ recorded states leave actions unchanged, with zero measured descending output throughout the intact rollouts. Recorded trajectories and state-matched images connect these findings to sustained movement, lateral drift, and termination events. The study characterizes an embodied recurrent control system whose tested locomotion is supported by direct body-and-command input and carried motor state, providing a concrete basis for subsequent comparisons of circuit structure and control resources.
\end{abstract}

\section{Introduction}
Locomotion is produced by a controller acting through a body in contact with its environment. The contribution of each component matters when asking how much neural computation a movement requires. Passive-dynamic bipeds and neuronal reflex controllers show how mechanics and local feedback can support coordinated walking \cite{collins,runbot}. Recurrent models of motor activity and embodied learning likewise motivate studying neural dynamics together with physical structure \cite{sussillo,gupta}. These observations raise a broader question about the relationship between neural resources and motor capability.

Insect connectomes offer a biologically grounded source of candidate network organization \cite{schmidgall}. Embodied fly models provide platforms for studying sensorimotor coupling \cite{neuromechfly,flybody}, and recent preprints examine connectome-based fly locomotion and robot navigation \cite{flygm,flynn}. Applying an insect-labelled recurrent core to a humanoid introduces a distinct interface problem: body observations and neural readouts must be coupled to a different joint arrangement and contact geometry.

We address the deployed-system question: \emph{how does a fly-inspired recurrent controller participate in humanoid locomotion, and which computational pathways support its behavior?} We study the existing T\_graph checkpoint in a simulated Unitree G1. Its core contains 3,609 continuous states, with a readout from 135 motor-neuron-labelled entries to 15 leg-and-waist targets. We formalize the neural--body loop, evaluate locomotion across specified terrain conditions, and use input/state interventions to distinguish the controller's nominal architecture from its effective control pathway.

The resulting evidence links three levels of description: model computation, body motion, and intervention response. The controller completes most of the evaluated conditions, while trajectory records expose lateral drift and specific failures. Clearing motor state causes early termination, whereas the tested depth and upstream-state substitutions leave actions unchanged. Together, these findings characterize a concrete cross-body control implementation and motivate controlled studies of the relationship between circuit structure, resource budget, and motor capability.

\section{Related work}
Compact locomotor controllers have a substantial history. Beyond passive mechanics \cite{collins}, examples include RunBot's reflex network with a six-neuron adaptive extension \cite{runbot}, six Matsuoka oscillators for a simulated five-link biped \cite{liucpg}, and an eight-neuron CPG combined with trajectory and joint mappings for a simulated quadruped \cite{eight}. The counts describe different components and must be interpreted alongside their surrounding control systems.

Humanoid controllers also span different resource scales. Steffen et al.\ use a reference H1 walking MLP with three 128-unit hidden layers before constructing a spiking approximation \cite{steffen}; Radosavovic et al.\ deploy a history-conditioned transformer for physical humanoid locomotion \cite{radosavovic}. Our 3,609-state count describes one recurrent core, with additional input and output computation. A useful efficiency comparison therefore requires matched tasks and total resource accounting.

Connectome-informed networks can preserve exact wiring, connectivity statistics, or circuit motifs \cite{schmidgall}. NeuroMechFly v2 and flybody examine embodied fly control \cite{neuromechfly,flybody}. FlyGM evaluates a fly-connectomic controller against graph and non-graph baselines, while FLYNN studies topology-based robot navigation and sensory loss \cite{flygm,flynn}. The present study focuses on an insect-labelled core operating through humanoid joint interfaces and on the effective pathways of a fixed deployed controller.

\section{Controller and neural--body coupling}
\subsection{Recurrent controller}
The supplied T\_graph package identifies its core as \texttt{ei\_graph} and attributes neuronal annotations to MaleCNS v1.0 \cite{malecns}. The annotations contain 1,286 descending-neuron-labelled entries (DN), 2,188 interneuron-labelled entries (IN), and 135 motor-neuron-labelled entries (MN). We use \emph{fly-inspired} to describe this annotated artificial recurrent representation. The available materials contain the deployed model and interfaces; the original biological adjacency, graph-extraction rules, and training configuration remain unavailable. Supplement~\ref{sec:provenance} specifies the provenance records.

The policy input $p_t\in\mathbb R^{82}$ contains projected gravity, angular velocity, a movement command, joint positions and velocities, and the previous raw action. A separate depth-processing branch with state $\beta_t\in\mathbb R^{256}$ produces a bottleneck $z_t\in\mathbb R^{128}$. Using row vectors, the motor computation is
\begin{align}
u_t &= [p_t,z_t]P^\top+b_P, \qquad u_t\in\mathbb{R}^{64},\label{eq:projection}\\
h_{t+1} &= (1-\alpha)\odot h_t+
 \alpha\odot\operatorname{ReLU}(h_tW+u_tB+b),\label{eq:core}\\
a_t &= h_{t+1}S_{\mathcal M}C^\top+b_C, \qquad a_t\in\mathbb{R}^{15},\label{eq:readout}
\end{align}
Here $h_t\in\mathbb R^{3609}$, $P\in\mathbb R^{64\times210}$, $W\in\mathbb R^{3609\times3609}$, and $B\in\mathbb R^{64\times3609}$. $S_{\mathcal M}$ selects the 135 MN-labelled entries and $C\in\mathbb R^{15\times135}$ maps them to actions. The update coefficient $\alpha$ ranges from 0.1473 to 0.7136. $W$ has 101,263 nonzero entries (0.777\% density); $B$ is dense with 230,976 nonzero entries. These are component counts, with states representing continuous artificial activations. Supplement~\ref{sec:details} gives the complete observation and execution equations.

\begin{figure}[tbp]
\centering\includegraphics[width=\linewidth]{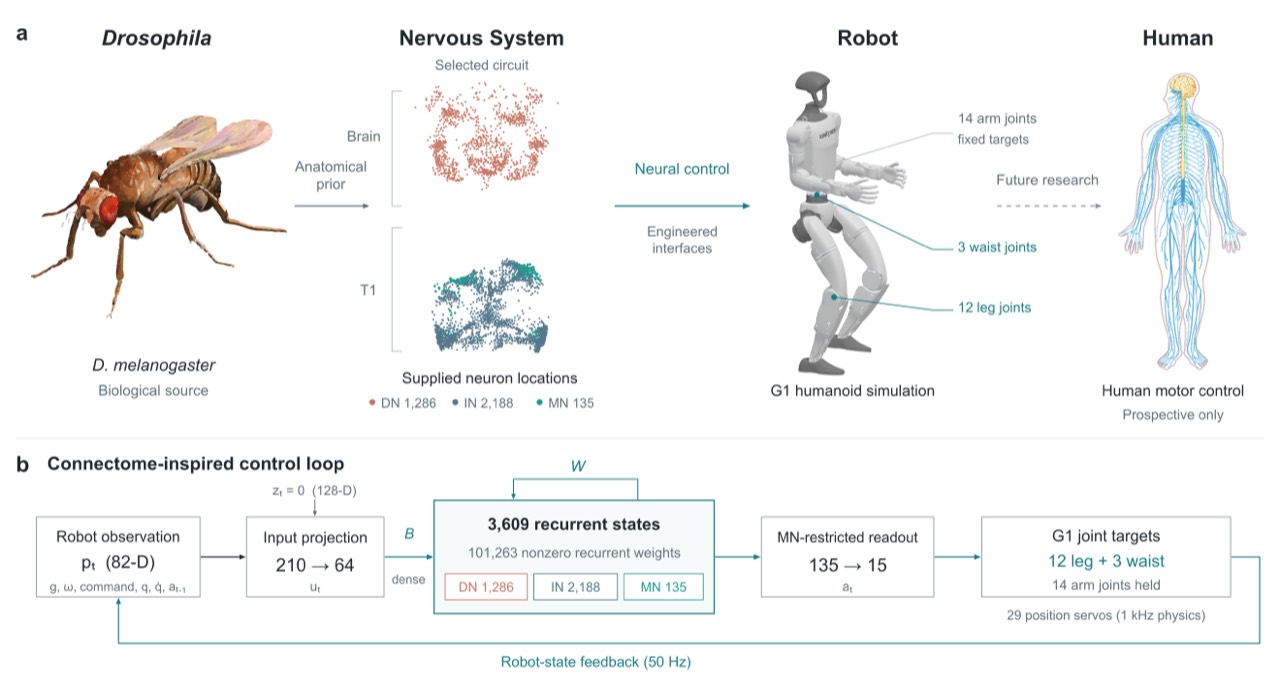}
\caption{Fly-inspired neural--body coupling. (a) Conceptual overview linking the biological source and supplied neuron locations to G1 simulation through engineered interfaces. The human motor-control panel denotes a prospective research direction. (b) The implemented control loop: robot observations pass through an input projection, the 3,609-state recurrent core, and an MN-restricted readout to generate 12 leg and three waist targets; 14 arm joints use fixed targets. Robot-state feedback runs at 50 Hz with 1 kHz physics. The 128-dimensional bottleneck $z_t$ is zero in the evaluated regime and is distinct from the 1,286 DN-labelled recurrent entries.}
\label{fig:architecture}
\end{figure}

\subsection{Coupling to the simulated body}
The G1 model has 29 actuated joints. The policy generates targets for 12 leg and three waist joints; the other 14 joints use fixed hold targets. Raw actions are clipped to $[-20,20]$, scaled by 0.25, added to per-joint offsets, and mapped to actuator order. Local position servos then act on the body. This is an engineered observation--action interface between the annotated neural representation and humanoid mechanics.

The policy runs at 50 Hz and MuJoCo \cite{mujoco} advances at 1,000 Hz, with each target held for twenty physics steps. Let $\mathcal O$ construct observations, $\mathcal N_\theta$ update neural states and output actions, $\mathcal A$ construct joint targets, and $\Phi_{0.001,\mathcal T}$ advance the body on terrain $\mathcal T$. The feedback loop is
\begin{equation}
\boxed{\begin{aligned}
p_t&=\mathcal O(x_t,c_t,a_{t-1}),\\
(h_{t+1},\beta_{t+1},a_t)&=\mathcal N_\theta(h_t,\beta_t,p_t,d_t),\\
q_t^{*,S}&=\mathcal A(a_t),\\
x_{t+1}&=\Phi_{0.001,\mathcal T}^{(20)}(x_t,q_t^{*,S}).
\end{aligned}}
\label{eq:coupling}
\end{equation}
where $x_t$ is the simulator state, $c_t$ the movement command, and $d_t$ the depth input; superscript $(20)$ denotes twenty successive physics steps. Body motion changes subsequent neural inputs, while neural outputs change subsequent body motion. The core, projections, readout, servos, and body jointly implement this loop.

\section{Evaluation}
\subsection{Terrain conditions and measures}
We evaluate seven fixed terrain instances: flat ground, 0--6 cm roughness, $10^\circ$ uphill and downhill slopes, 10 cm ascending and descending stairs, and 10 cm discrete steps. Each intact controller is tested at forward commands $v\in\{0.3,0.5,0.8\}$ m/s and initial yaw offsets $\{-0.05,0,0.05\}$ rad, giving 63 conditions. Each run starts from the exported spawn pose with zero recurrent states and lasts up to 12 s (600 policy calls). The terrain instances' relation to training data is unknown.

Termination occurs if ground-relative pelvis height is below 0.4 m, absolute roll or pitch exceeds 1 rad, or ground-relative height is nonfinite. The latter includes leaving the rough terrain's valid height-query domain. Success requires survival for 12 s and forward displacement $\Delta x\geq0.7vT$, where $T=12$ s. This measures survival and forward progress; lateral deviation and termination type are reported separately. Velocity RMSE is measured over non-terminated observations, so failed runs have shorter measurement windows.

T\_graph receives a fixed normalized-zero depth array throughout. The supplied R1 reference receives the same 82-dimensional body-and-command vector plus 187 privileged terrain-height samples. R1 provides context under the same physical scenarios; its sensory access and undocumented training budget differ from T\_graph. Both are fixed inference checkpoints.

\subsection{State and input interventions}
We run 21 additional T\_graph episodes at nominal yaw, setting $h_t=0$ before every policy call. This intervention jointly removes carried state and the recurrent interaction term in Equation~\ref{eq:core}. The full study comprises 63 intact T\_graph, 63 R1, and 21 reset episodes.

At every 50th step from step zero in the intact nominal-yaw graph trajectories, we also save inputs and perform six independent substitutions: constant depth $-0.5$, constant depth $+0.5$, alternating $\pm0.5$ depth, zero upstream state, zero motor state, and zero body-and-command input. The last removes all 82 input components, including commands and previous actions. These 1,512 counterfactual calls at \ProbeCount\ recorded states leave the original rollouts unchanged. For input state $\xi_t=(p_t,d_t,\beta_t,h_t)$ and substitution $\mathcal I$, action sensitivity is
\begin{equation}
\Delta_{\mathcal I}(t)=\left\|\pi_\theta(\mathcal I(\xi_t))-\pi_\theta(\xi_t)\right\|_\infty,
\label{eq:intervention}
\end{equation}
where $\pi_\theta$ is the raw-action component of $\mathcal N_\theta$.

We retain every planned outcome. Eleven exact replays of existing episodes supply 115 archived states and 28 annotated images. The images preserve recorded poses and terrain geometry; they visualize the evaluated behavior rather than add independent trials. Runtime, complete outcomes, and rendering details are provided in the supplement.

\section{Results}
\subsection{Locomotion across the evaluated terrains}
\begin{figure}[tbp]
\centering\includegraphics[width=\linewidth]{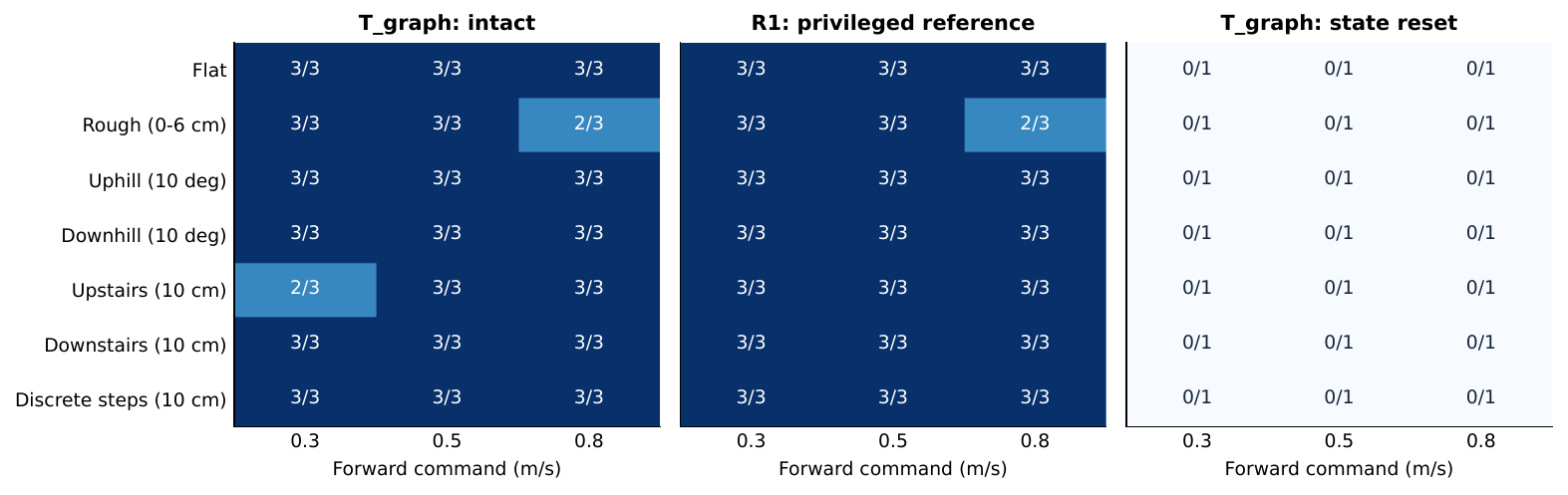}
\caption{All planned condition outcomes, including every failed condition. Numbers are success counts divided by the number of initial-yaw conditions (three for intact checkpoints, one for the state-reset intervention).}
\label{fig:outcomes}
\end{figure}

T\_graph completes \GraphSuccess/63 conditions, compared with \TeacherSuccess/63 for R1 (Figure~\ref{fig:outcomes}). The graph controller has one termination under the ground-height rule and one upright run with insufficient forward progress. On flat ground, both slopes, descending stairs, and discrete steps, it meets the stated criterion in all nine speed--yaw combinations. The two unsuccessful conditions occur on rough terrain and ascending stairs. Supplement~\ref{sec:conditions} reports complete terrain--speed summaries.

\begin{figure}[tbp]
\centering\shot{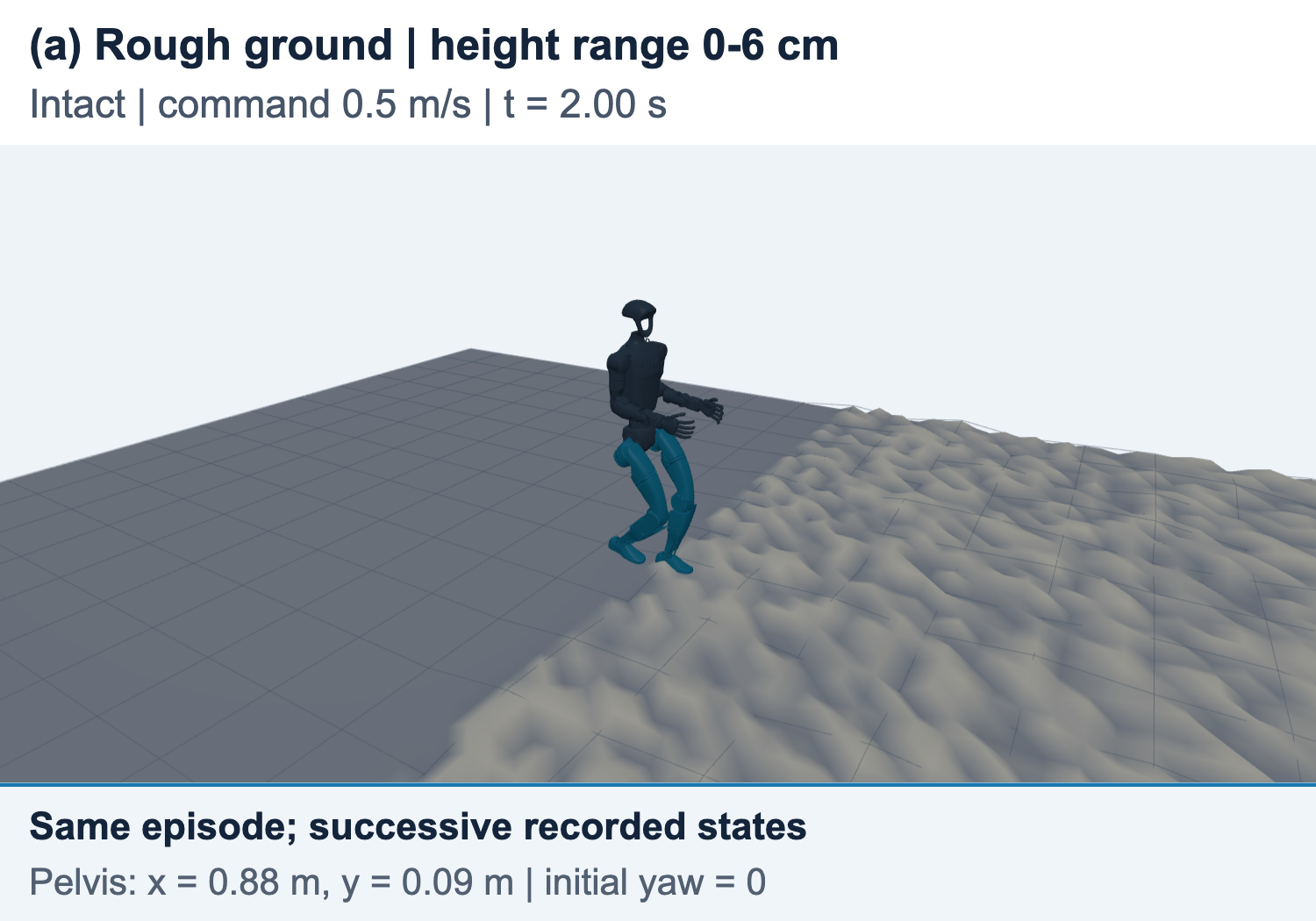}\hfill\shot{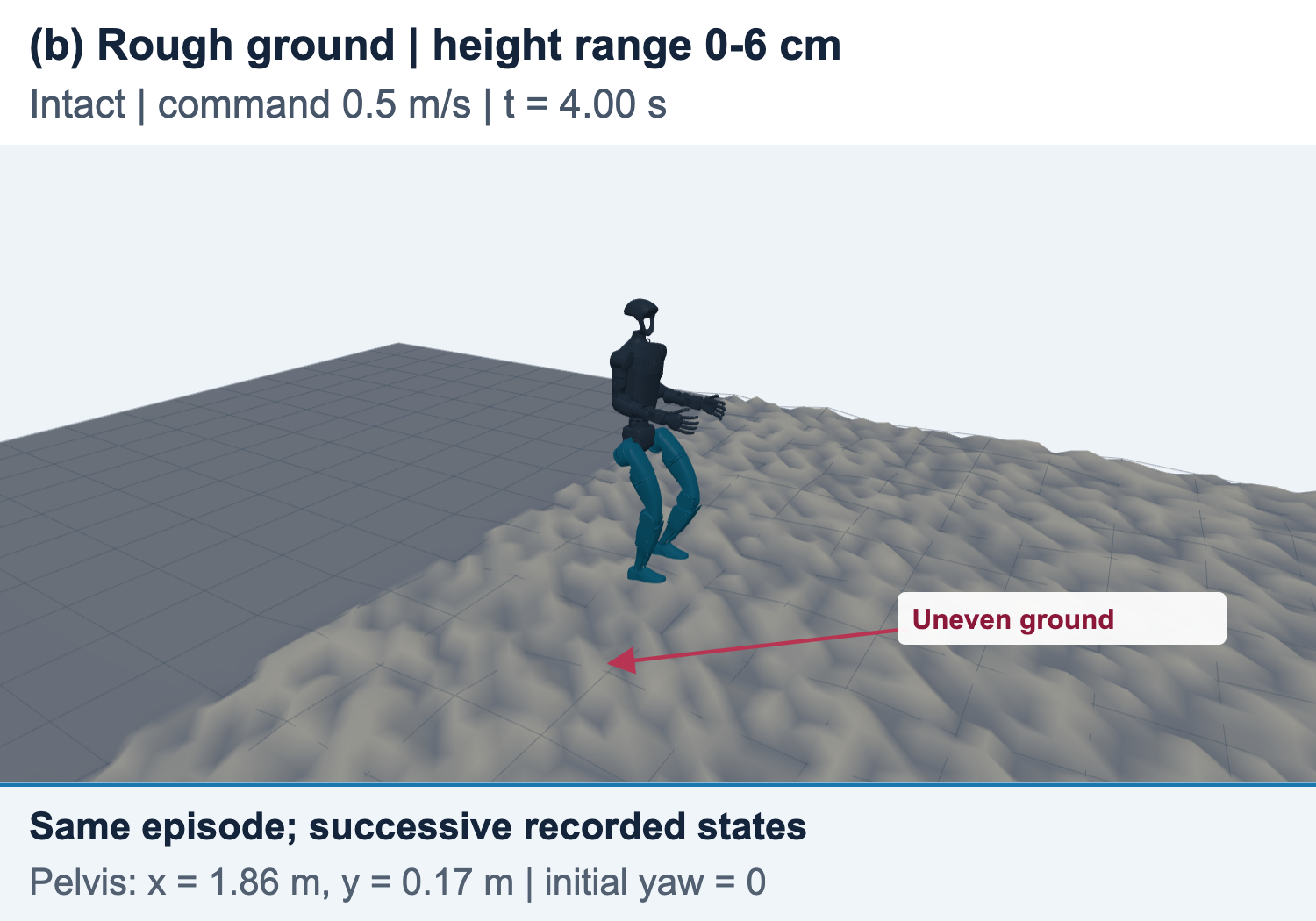}\par\vspace{4pt}
\shot{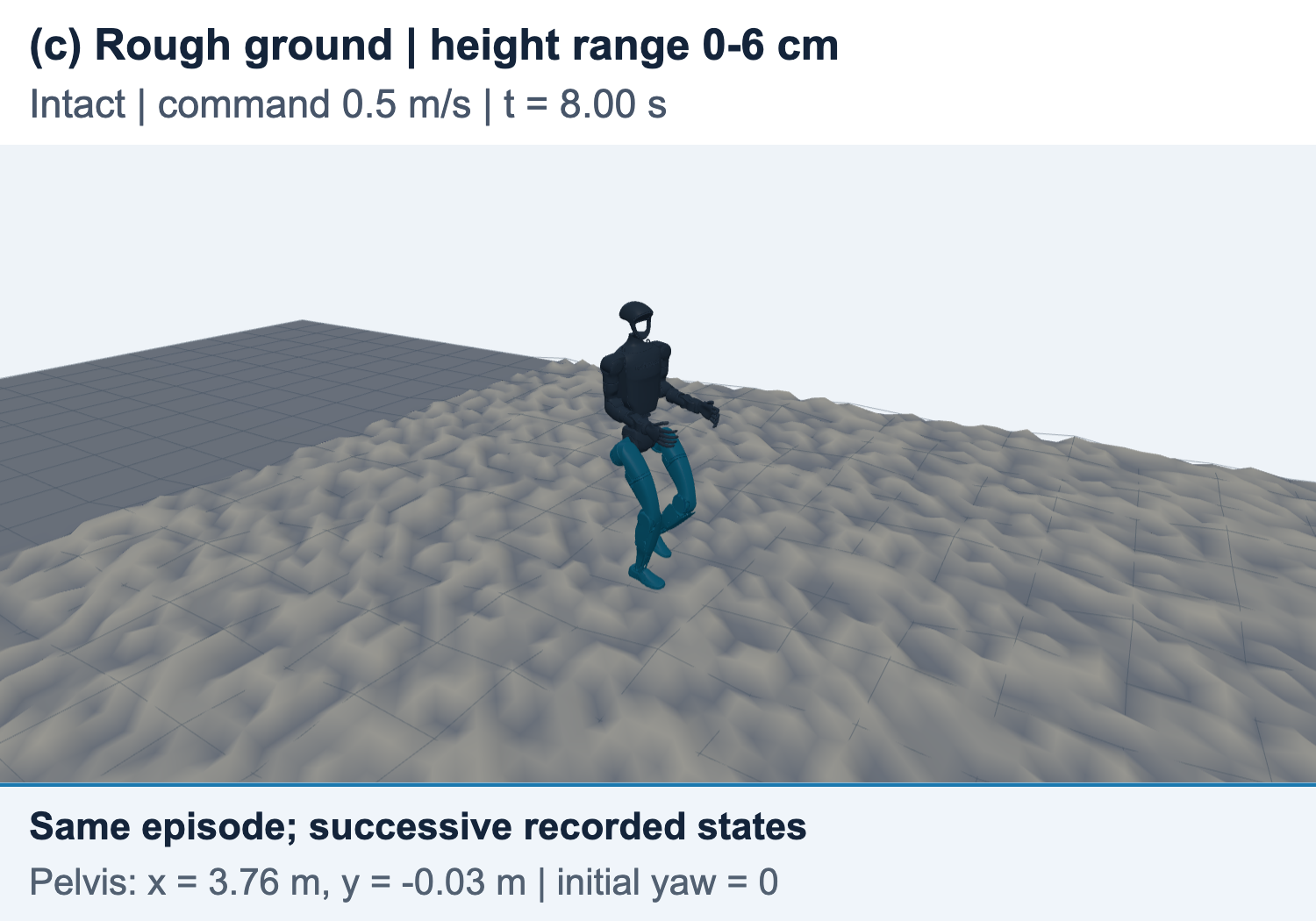}\hfill\shot{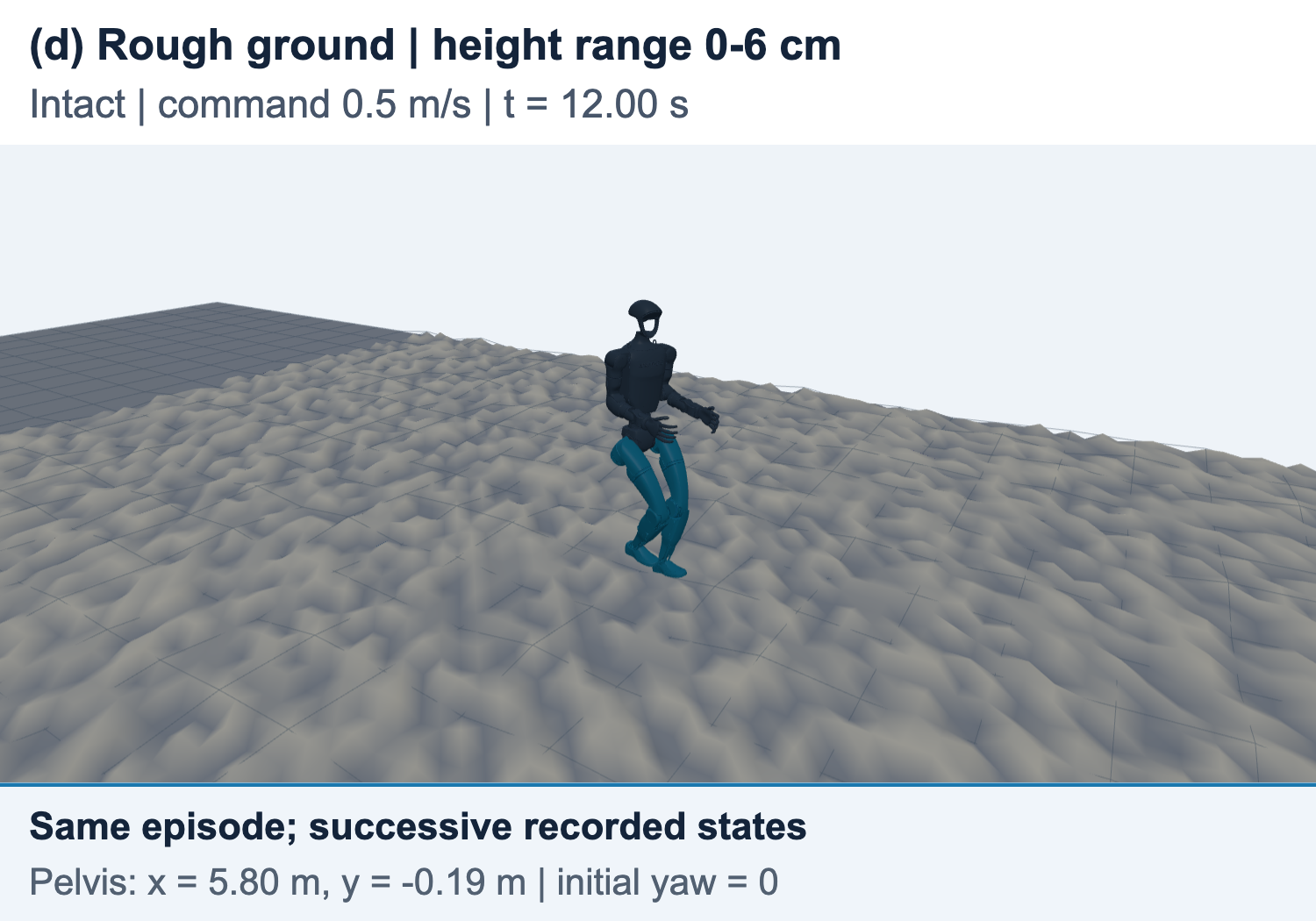}
\caption{One rough-ground rollout at a constant 0.5 m/s forward command and zero initial yaw, shown at 2, 4, 8, and 12 s. The arrow identifies uneven ground near the robot. All poses come from the same intact T\_graph replay with synthetic zero depth, documenting the progression of body posture and position over the recorded episode.}
\label{fig:rough-seq}
\end{figure}

The recorded motion links these outcomes to terrain interaction. The 0.5 m/s rough-ground sequence (Figure~\ref{fig:rough-seq}) shows sustained movement through the 12 s horizon. The supplemental atlas and slope/stair sequences show the same checkpoint under common command and time settings. These images establish the pose and displacement attained in each recorded episode.

Forward progress and path control separate in some cases (Figure~\ref{fig:failure-images}). At 0.8 m/s on discrete steps with nominal yaw, a successful run ends 3.17 m laterally from its start, outside the nominal 6 m-wide obstacle corridor. At 0.3 m/s on ascending stairs, the nominal-yaw run remains upright but advances 1.77 m, below the required 2.52 m. The nominal-yaw 0.8 m/s rough run reaches $y=-3.01$ m and terminates at 11.70 s when its height query leaves the valid domain. We identify this as a boundary termination; the record alone does not establish physical toppling.

\subsection{Carried motor state supports the deployed behavior}
\begin{figure}[tbp]
\centering\shot{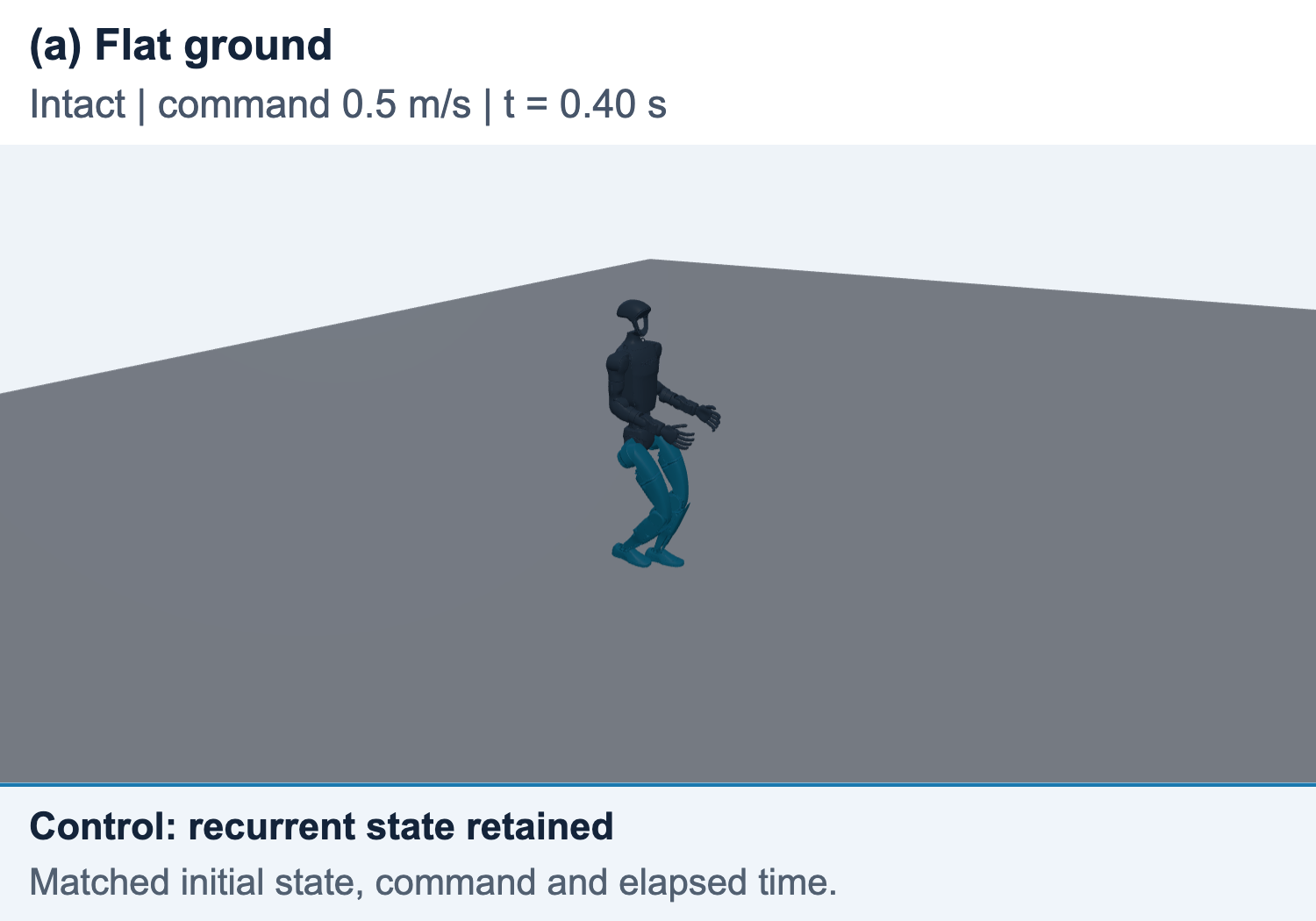}\hfill\shot{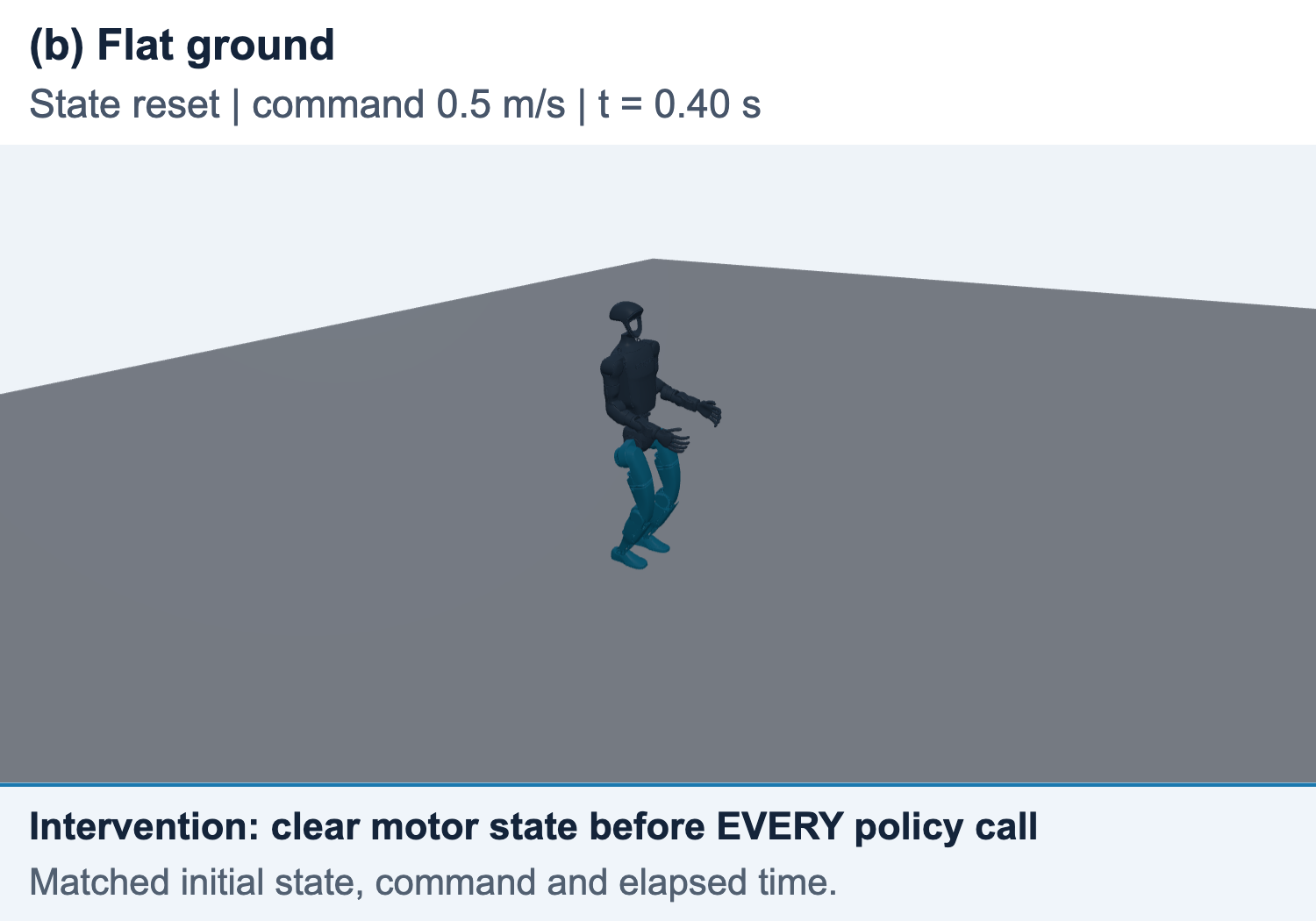}\par\vspace{4pt}
\shot{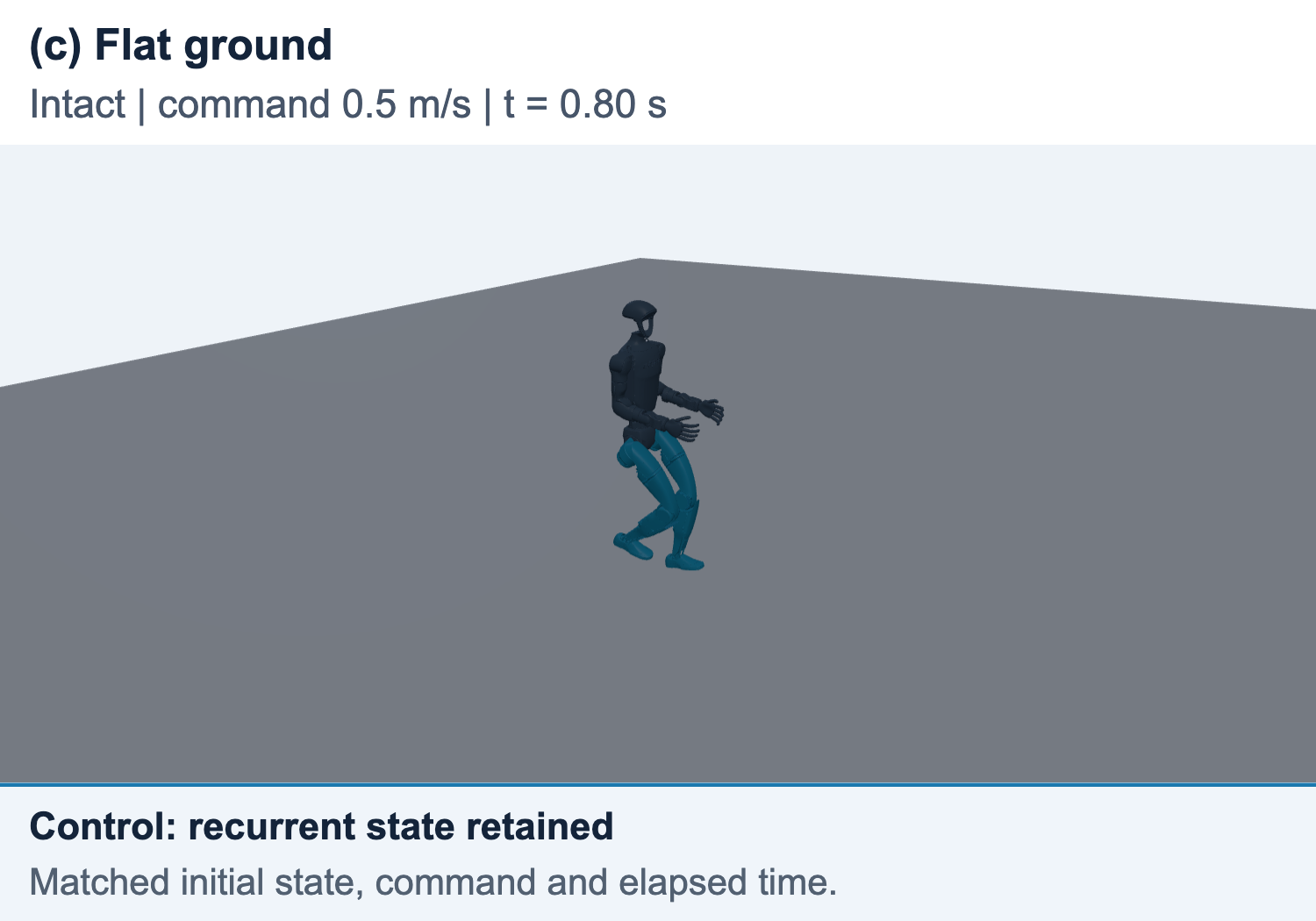}\hfill\shot{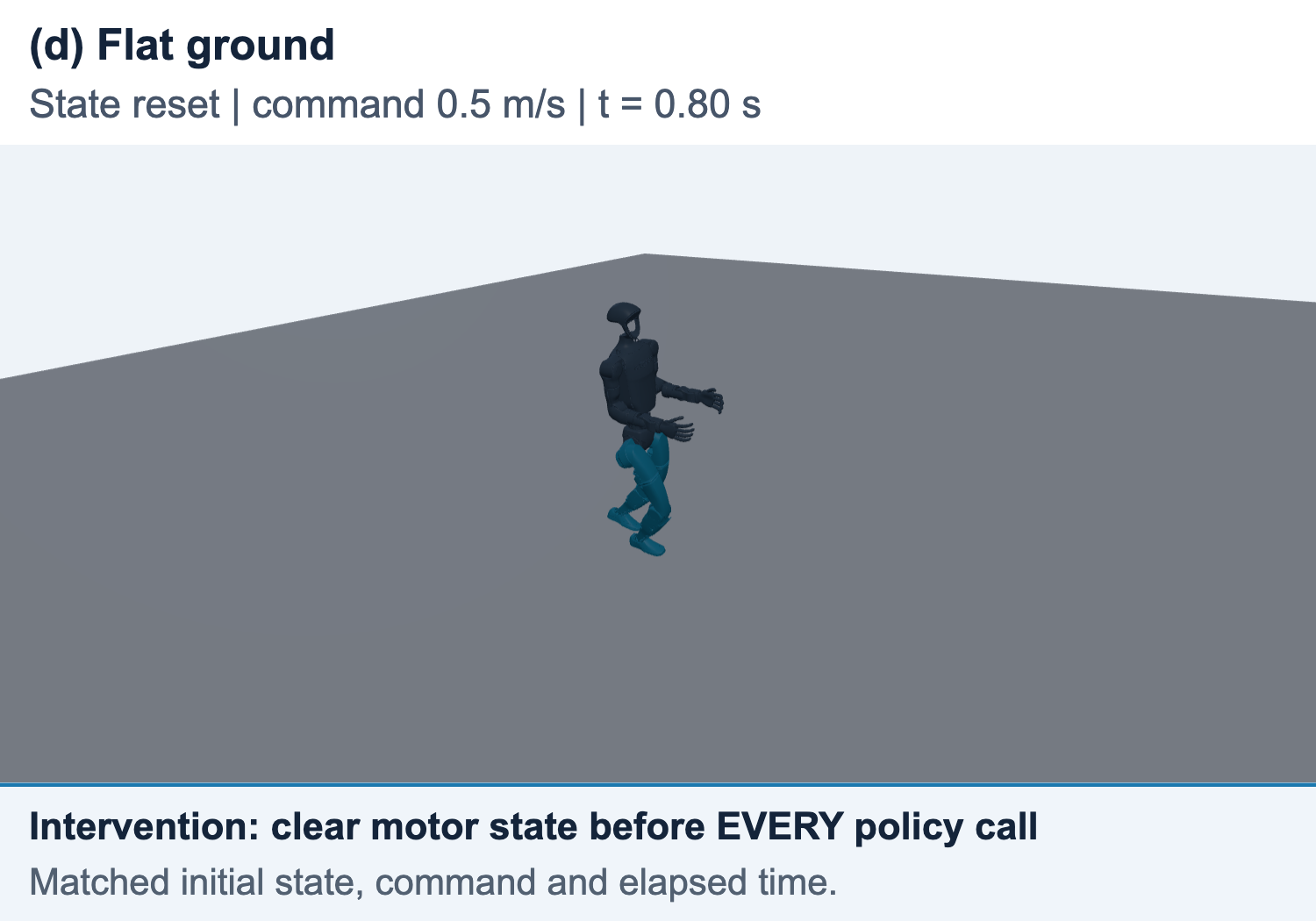}\par\vspace{4pt}
\shot{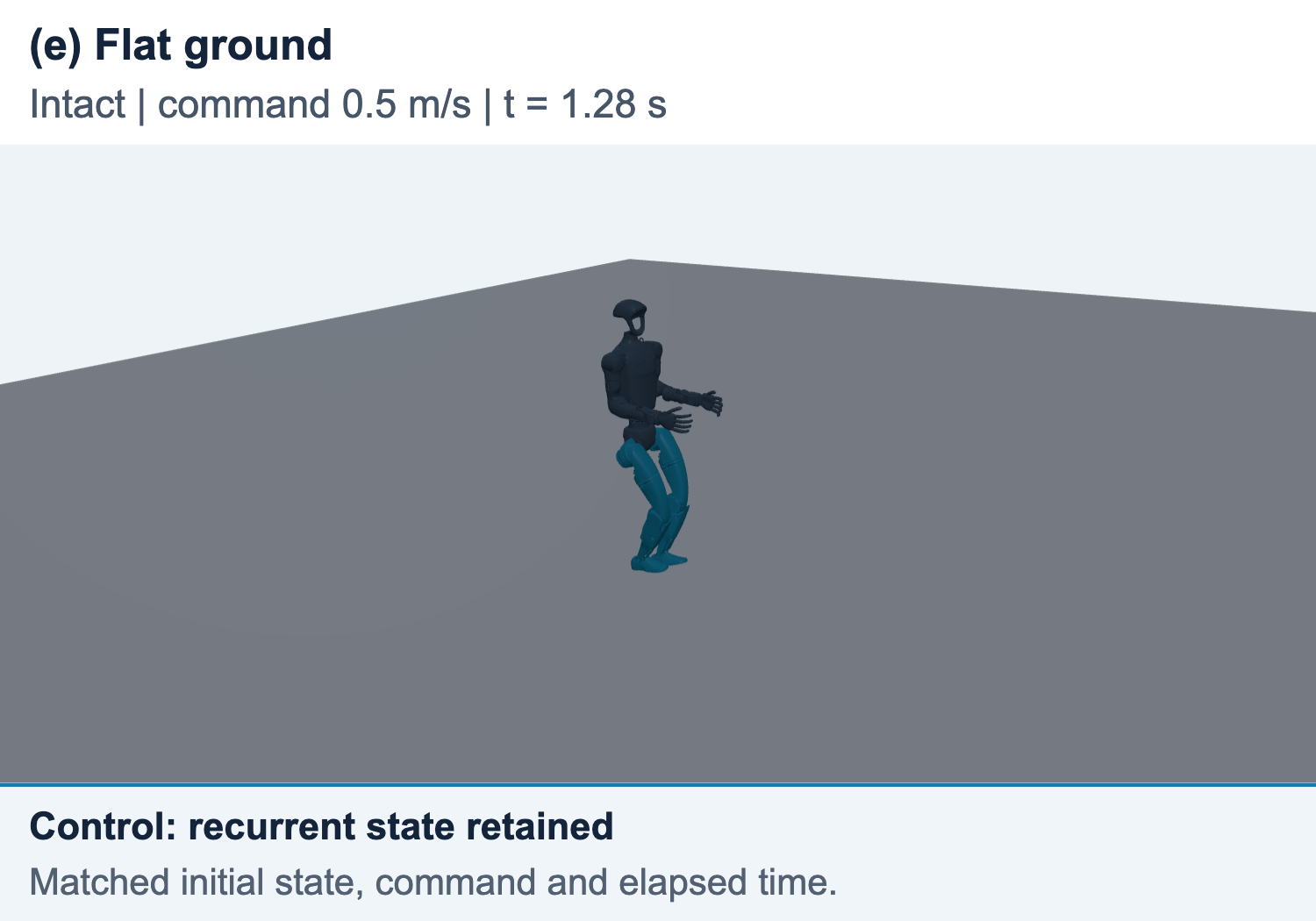}\hfill\shot{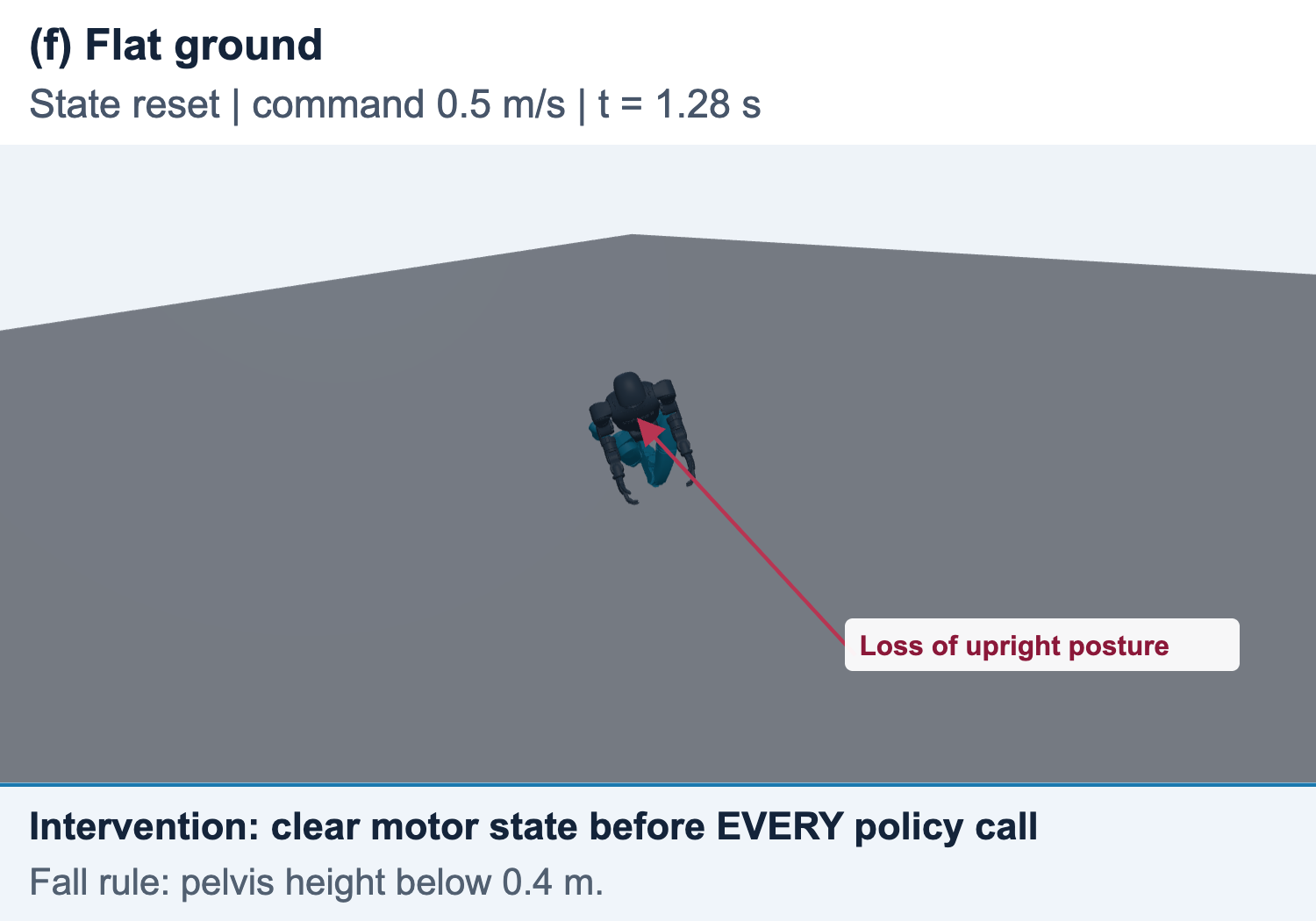}
\caption{Matched-time state-reset intervention on flat ground at 0.5 m/s. Left: intact state; right: $h_t=0$ before every policy evaluation. Rows show 0.40, 0.80, and 1.28 s. The final right panel reaches the fall rule while the intact counterpart remains upright. The intervention jointly removes carried state and recurrent interaction within the same fixed checkpoint. Panel letters identify each captured image; columns and rows specify the comparison.}
\label{fig:reset-images}
\end{figure}

At nominal yaw, intact T\_graph succeeds in \GraphNominalSuccess/21 conditions; resetting its motor state yields \ResetSuccess/21 and terminates every run between \ResetFallMin\ and \ResetFallMax\ s (median \ResetFallMedian\ s). The matched flat-ground sequence in Figure~\ref{fig:reset-images} links this intervention to loss of upright posture: at 1.28 s the reset run reaches the pelvis-height threshold while the intact counterpart remains upright.

One-step motor-state substitutions have median action sensitivity \HVDeltaMedian\ over noninitial probes and maximum \HVDeltaMax\ across all probes, using Equation~\ref{eq:intervention}. The action and rollout results consistently identify dependence on the carried motor-state mechanism of this checkpoint. The intervention acts on the complete recurrent state, so the result concerns system-level state dependence.

\subsection{The effective input pathway bypasses the depth branch}
\begin{figure}[tbp]
\centering\includegraphics[width=\linewidth]{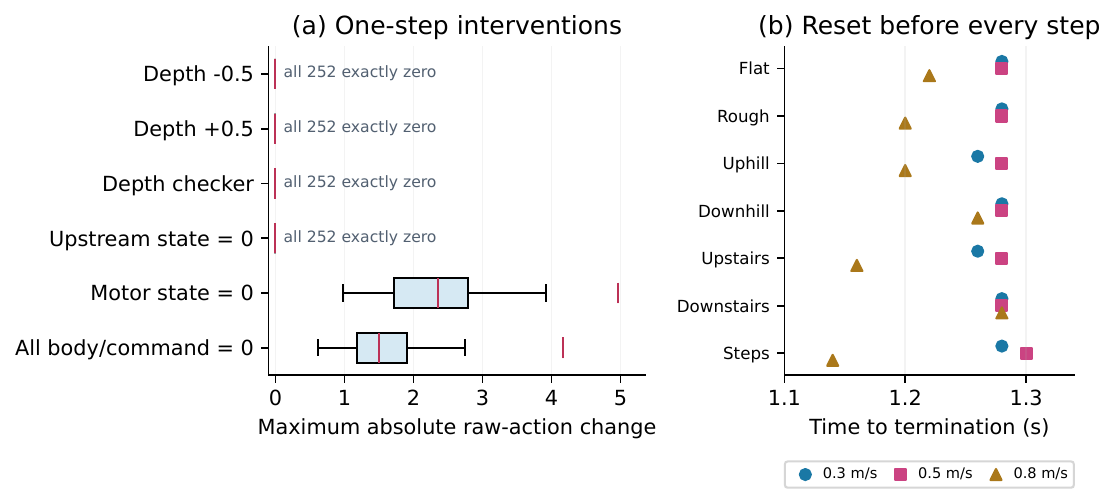}
\caption{What changes the deployed controller? (a) Raw-action differences under six one-step substitutions at all 252 sampled states, including the 21 initial states. Boxes show median and interquartile range; whiskers use 1.5 interquartile ranges; red vertical markers show maxima. All tested depth and upstream-state substitutions give exactly zero action difference. (b) All 21 closed-loop motor-state-reset episodes terminate between 1.14 and 1.30 s. Distributions summarize sampled states and specified conditions of one graph checkpoint.}
\label{fig:interventions}
\end{figure}

Across \GraphSteps\ intact graph-policy calls, the logged descending output $z_t$ is zero. At all \ProbeCount\ recorded states, each tested depth substitution gives zero raw-action difference; zeroing the upstream recurrent state also gives zero difference. In contrast, zeroing the body-and-command vector produces a median maximum-coordinate difference of \PropDeltaMedian. Figure~\ref{fig:interventions} summarizes the sensitivities and reset termination times.

These observations locate the effective tested pathway in direct body-and-command input and recurrent motor state. DN-labelled entries inside the core can still be active: they receive the dense engineered input and participate in recurrence, whereas $z_t$ is a separate upstream bottleneck. Recorded group activations are shown in Supplement~\ref{sec:activity}. The origin of the inactive upstream branch remains a training-history question.

\FloatBarrier
\section{Discussion and conclusion}
The study connects an explicit recurrent computation to humanoid behavior and to interventions on its deployed pathways. The engineering interface is central: it translates body observations into core drive and selected core states into leg-and-waist targets. Its cross-body significance is a functional coupling between an insect-labelled representation and humanoid mechanics; the MN-labelled linear readout defines an engineered action map, without establishing biological homology between insect and humanoid joints.

The results also distinguish recurrent state dependence from the broader value of a connectome-derived architecture. A controller that relies on carried state can fail when that state is repeatedly erased. Establishing an advantage of biological wiring or a minimum sufficient network size requires separately trained, resource-matched controls and scale comparisons. The 3,609-state count is a characterization of this core, not a demonstrated resource advantage over conventional policies.

Evidence here concerns a fixed checkpoint, seven fixed terrain instances, and controlled initial yaw offsets under synthetic zero depth. The success criterion allows lateral drift, and the reference has privileged sensing. Graph extraction and training records remain unavailable in the supplied materials. Simulated contact settings and disabled self-collisions further define the physical test conditions. These boundaries delimit the present characterization; broader perturbation tests, matched architecture comparisons, and supervised hardware validation are subsequent steps.

Within that scope, the study establishes three connected observations: the deployed controller sustains locomotion in \GraphSuccess/63 specified conditions; recurrent motor state is consequential for that behavior; and the measured depth-to-descending branch is inactive while direct body input remains effective. This links neural computation, terrain-dependent motion, and pathway intervention in one inspectable humanoid control system.

\label{maintext-end}
\section*{Data availability and AI assistance}
The author-held archive contains the protocol, 147 episode summaries, sampled trajectories, intervention records, analysis scripts, model hashes, and state-matched visualization records. The original inference bundle is a separate permission-dependent dependency; a public archive identifier has yet to be established. Full evaluation reruns require that bundle. OpenAI Codex assisted with software inspection, evaluation implementation and execution, analysis, figures, and writing. Human authors are responsible for scientific verification and interpretation. All reported experiments are simulations and software inspections.

\clearpage
\begingroup\small\setlength{\parskip}{2pt}

\input{references.tex}
\endgroup
\clearpage
\input{supplement.tex}
\end{document}

%% file: results.tex
\newcommand{\GraphSuccess}{61}
\newcommand{\TeacherSuccess}{62}
\newcommand{\ResetSuccess}{0}

\newcommand{\GraphNominalSuccess}{19}
\newcommand{\ProbeCount}{252}
\newcommand{\GraphSteps}{37785}
\newcommand{\ResetFallMin}{1.14}
\newcommand{\ResetFallMax}{1.30}
\newcommand{\ResetFallMedian}{1.28}

\newcommand{\HVDeltaMedian}{2.436}
\newcommand{\HVDeltaMax}{4.965}
\newcommand{\PropDeltaMedian}{1.507}

\newcommand{\ConditionRows}{%
Flat & 0.3 & 3/3 & 4.85 & 0.139 & 3/3 \\
Flat & 0.5 & 3/3 & 6.85 & 0.120 & 3/3 \\
Flat & 0.8 & 3/3 & 9.59 & 0.125 & 3/3 \\
Rough, 0--6 cm & 0.3 & 3/3 & 3.75 & 0.102 & 3/3 \\
Rough, 0--6 cm & 0.5 & 3/3 & 6.02 & 0.119 & 3/3 \\
Rough, 0--6 cm & 0.8 & 2/3 & 7.92 & 0.271 & 2/3 \\
Uphill, $10^\circ$ & 0.3 & 3/3 & 3.64 & 0.066 & 3/3 \\
Uphill, $10^\circ$ & 0.5 & 3/3 & 5.40 & 0.089 & 3/3 \\
Uphill, $10^\circ$ & 0.8 & 3/3 & 8.30 & 0.163 & 3/3 \\
Downhill, $10^\circ$ & 0.3 & 3/3 & 4.46 & 0.120 & 3/3 \\
Downhill, $10^\circ$ & 0.5 & 3/3 & 6.09 & 0.122 & 3/3 \\
Downhill, $10^\circ$ & 0.8 & 3/3 & 8.77 & 0.196 & 3/3 \\
Upstairs, 10 cm & 0.3 & 2/3 & 2.70 & 0.132 & 3/3 \\
Upstairs, 10 cm & 0.5 & 3/3 & 5.39 & 0.142 & 3/3 \\
Upstairs, 10 cm & 0.8 & 3/3 & 8.39 & 0.184 & 3/3 \\
Downstairs, 10 cm & 0.3 & 3/3 & 4.75 & 0.160 & 3/3 \\
Downstairs, 10 cm & 0.5 & 3/3 & 6.32 & 0.160 & 3/3 \\
Downstairs, 10 cm & 0.8 & 3/3 & 9.52 & 0.176 & 3/3 \\
Discrete steps, 10 cm & 0.3 & 3/3 & 3.74 & 0.103 & 3/3 \\
Discrete steps, 10 cm & 0.5 & 3/3 & 5.68 & 0.167 & 3/3 \\
Discrete steps, 10 cm & 0.8 & 3/3 & 7.65 & 0.206 & 3/3 \\
}

%% file: supplement.tex
\appendix
\renewcommand{\thesection}{S\arabic{section}}
\setcounter{section}{0}
\setcounter{figure}{0}
\setcounter{table}{0}
\setcounter{equation}{0}
\renewcommand{\thefigure}{S\arabic{figure}}
\renewcommand{\thetable}{S\arabic{table}}
\renewcommand{\theequation}{S\arabic{equation}}
\renewcommand{\theHfigure}{supp.\arabic{figure}}
\renewcommand{\theHtable}{supp.\arabic{table}}
\renewcommand{\theHequation}{supp.\arabic{equation}}
\fancyhead[R]{\small Supplement}
\renewcommand{\shot}[1]{\includegraphics[width=.44\linewidth]{screenshots/#1.png}}
\section*{Supplementary material}
\section{Complete interfaces and execution}\label{sec:details}
\subsection{Observation and upstream branch}
All vectors are row vectors. The 82 observation components are projected gravity (3), body angular velocity (3), commanded planar and yaw velocity (3), relative joint position (29), joint velocity (29), and previous raw action (15). Superscripts $I$ and $S$ denote Isaac and SDK order, respectively.
\begin{align}
r_t&=[\,g_t^B,\ \omega_t^B,\ c_t,\ q_t^I-q_0^I,\ \dot q_t^I,\ a_{t-1}\,]\in\mathbb R^{82},\label{eq:obs-raw}\\
p_t&=\operatorname{clip}_{[-100,100]}(r_t)D,\label{eq:obs-scale}\\
D&=\operatorname{diag}(\mathbf1_3,0.2\mathbf1_3,\mathbf1_3,\mathbf1_{29},0.05\mathbf1_{29},\mathbf1_{15}).\nonumber
\end{align}
Here $g_t^B$ is body-frame unit gravity and $\omega_t^B$ is angular velocity. The command vector contains planar velocity and yaw-rate targets; $q_0^I$ is the exported default pose. There is no direct base linear-velocity observation. The previous action is retained before actuator-target clipping. Initial actions and recurrent states are zero.

The upstream branch compresses a $36\times32$ depth array into 32 features, combines them with $p_t$, and updates $\beta_t\in\mathbb R^{256}$:
\begin{align}
\beta_{t+1}&=\mathcal G_{\theta}([\mathcal E_d(d_t),p_t],\beta_t),\label{eq:upstream}\\
z_t&=\operatorname{ReLU}(\beta_{t+1}Q^\top+b_z)\in\mathbb R^{128}.\label{eq:descending}
\end{align}
$\mathcal E_d$ is the depth encoder and $\mathcal G_\theta$ its downstream processing and state update. The normalized-zero depth input used here is synthetic. The 128-dimensional bottleneck is distinct from the 1,286 DN-labelled entries in the motor core. For a target core state $j$, the update in main-text Equation~\ref{eq:core} expands to
\begin{equation}
h_{t+1,j}=(1-\alpha_j)h_{t,j}+\alpha_j\left[\sum_{i=1}^{3609}h_{t,i}W_{ij}+\sum_{\ell=1}^{64}u_{t,\ell}B_{\ell j}+b_j\right]_+.
\label{eq:cell}
\end{equation}
The first sum mixes recurrent states and the second injects encoded body information. The same update acts on DN-, IN-, and MN-labelled entries through the dense input projection.

\subsection{Action targets, servos, and physics}
\begin{figure}[htbp]
\centering\includegraphics[width=\linewidth]{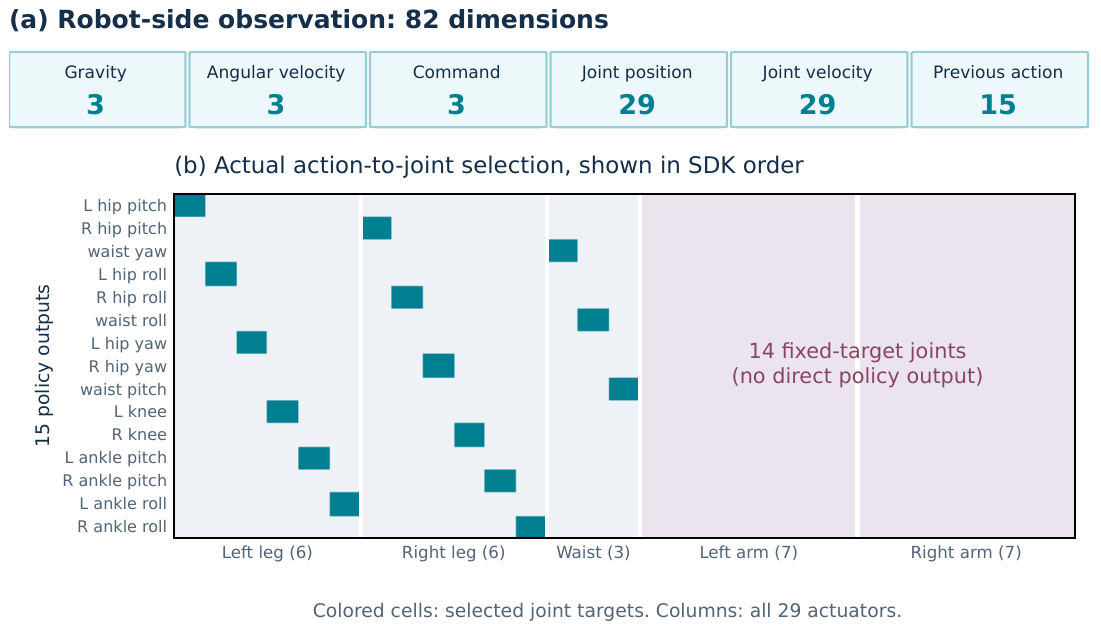}
\caption{What is connected across embodiments? (a) The six robot-side observation groups total 82 dimensions. (b) The deployed action selection and order map: 15 outputs set six left-leg, six right-leg, and three waist targets; the remaining 14 joints use fixed targets. Rows follow policy action order and columns follow actuator order. This engineered interface translates the neural readout into robot joint targets.}
\label{fig:interface}
\end{figure}

Let $E\in\{0,1\}^{29\times15}$ select the policy-controlled Isaac-order joints, with $E^\top E=I_{15}$, and let $\Pi\in\{0,1\}^{29\times29}$ map Isaac indices to SDK indices. The exact target construction is
\begin{align}
v_t&=o+0.25\operatorname{clip}_{[-20,20]}(a_t)\in\mathbb R^{15},\label{eq:target15}\\
q_t^{*,I}&=q_{\mathrm{hold}}^I(I_{29}-EE^\top)+v_tE^\top,\label{eq:target29}\\
q_t^{*,S}&=q_t^{*,I}\Pi^\top.\label{eq:permutation}
\end{align}
$o$ contains the 15 exported offsets, and $q_{\mathrm{hold}}^I$ gives fixed targets for unselected joints. Figure~\ref{fig:interface} shows the selection. The unit-gear affine servo law at physical substep $s$ is
\begin{equation}
\tau_{t,s}=\operatorname{clip}_{[-\tau_{\max},\tau_{\max}]}\!\left(k_p\odot(q_t^{*,S}-q_{t,s}^S)-k_d\odot\dot q_{t,s}^S\right).
\label{eq:servo}
\end{equation}
with gains and effort limits from the deployment contract. MuJoCo evaluates this actuator law with the model's other forces and contacts. Defining one physical step by $\Phi$, targets are held for twenty substeps:
\begin{equation}
x_{t,0}=x_t,\quad x_{t,s+1}=\Phi_{0.001,\mathcal T}(x_{t,s},q_t^{*,S}),\quad
s=0,\ldots,19,\quad x_{t+1}=x_{t,20}.
\label{eq:physics}
\end{equation}
The body has 29 actuated coordinates and a simulated mass of approximately 34.341 kg after the package's 1 kg pelvis adjustment. The contract selects \texttt{implicitfast}, disables self-collisions, and sets \texttt{solref} to $(0.005,1)$. Its empty \texttt{solimp} array leaves model impedance settings unchanged. Both evaluated controllers use the same inspected physics settings.

\begin{figure}[htbp]
\centering\includegraphics[width=\linewidth]{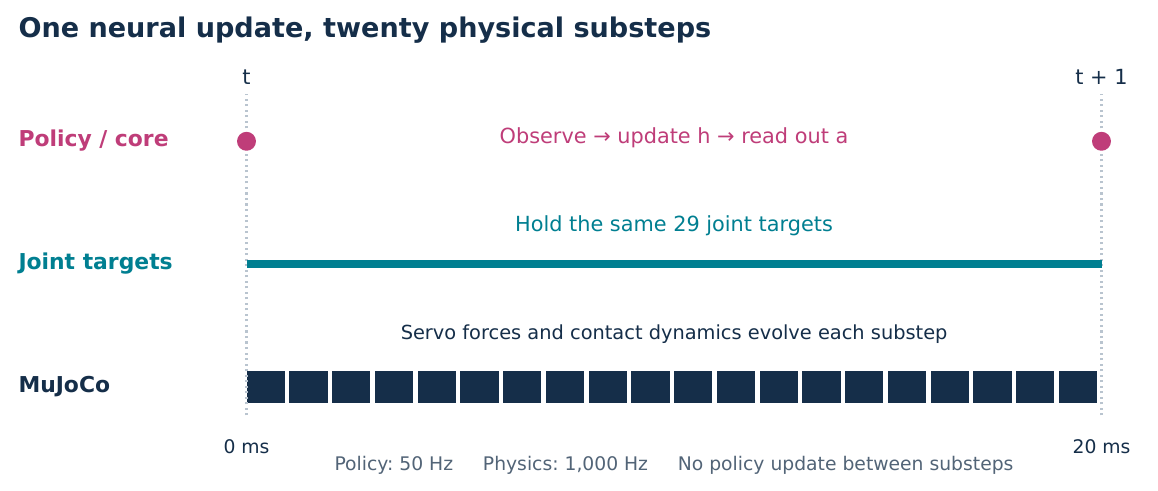}
\caption{Temporal coupling of neural and body dynamics. A policy call generates one set of joint targets; MuJoCo advances twenty 1 ms substeps under those targets before the next observation and neural update. Joint targets remain fixed during the interval, while servo forces respond to changing positions and velocities.}
\label{fig:timing}
\end{figure}
\FloatBarrier

\section{Core organization and resource context}
The annotations assign 1,286 entries to BRAIN and 2,323 to T1, organized into 37 visualization groups. The exported recurrent matrix has 101,263 nonzero entries, including two diagonal entries. All outgoing nonzero weights within a presynaptic row share a sign. The 1,979 positive and 1,544 negative annotations agree with these signs; 86 entries have unspecified annotation sign and contribute 1,452 nonzero outgoing weights. Each counted matrix entry is one weighted model connection, distinct from biological synapse multiplicity.

\begin{figure}[htbp]
\centering\includegraphics[width=\linewidth]{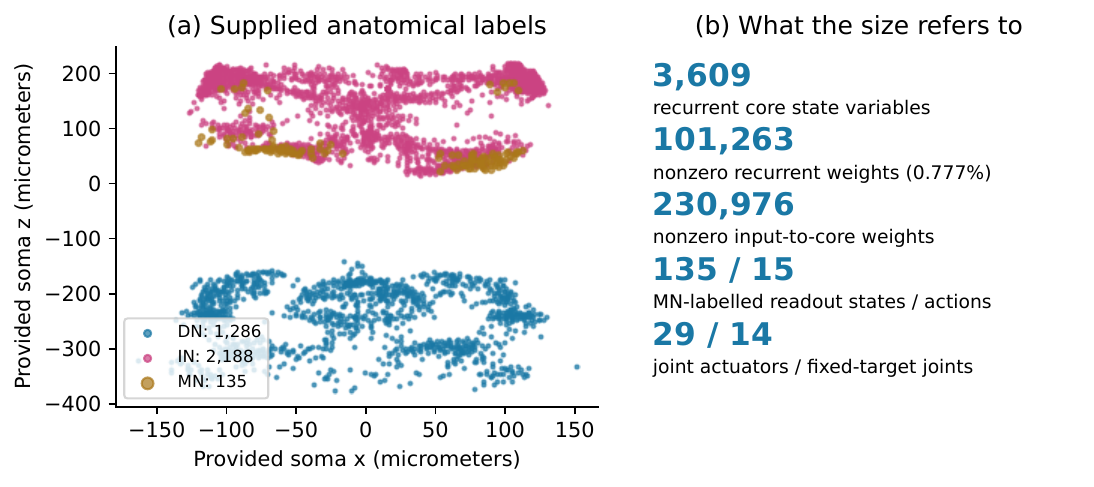}
\caption{The bounded core and its surrounding control resources. (a) Projection of supplied soma coordinates, colored by annotation class. (b) Component-level counts for recurrent states, nonzero recurrent weights, dense input weights, readout states, and actuators. The 15 policy-controlled joints and 14 fixed-target joints together form the 29-actuator body.}
\label{fig:core-size}
\end{figure}

\begin{figure}[htbp]
\centering\includegraphics[width=\linewidth]{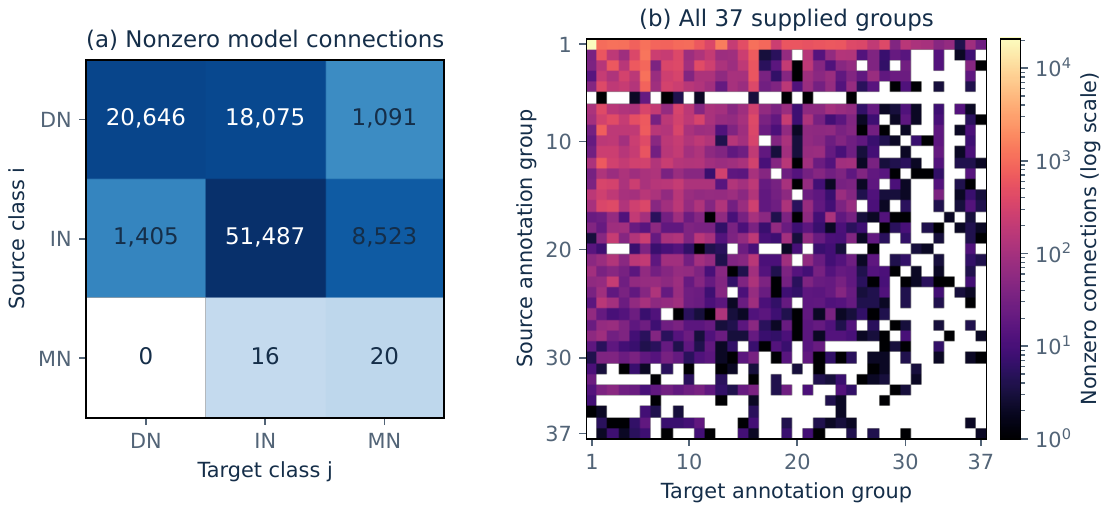}
\caption{Connectivity computed from the deployed recurrent matrix. (a) Nonzero weights aggregated by source and target annotation class; the nine counts sum to 101,263. (b) The same support aggregated into 37 supplied groups. Positive counts use logarithmic color scaling; empty cells indicate zero. The plots describe weighted model-connection counts and their distribution across annotations.}
\label{fig:connectivity}
\end{figure}

For supplied annotation groups $\mathcal G_a$, the inspected support and grouped counts are
\begin{equation}
A^{\mathrm{exp}}_{ij}=\mathbf{1}[W_{ij}\ne0],\qquad
N_{ab}=\sum_{i\in\mathcal G_a}\sum_{j\in\mathcal G_b}A^{\mathrm{exp}}_{ij}.
\label{eq:graph-support}
\end{equation}
where $W_{ij}$ couples source $i$ to target $j$. Figures~\ref{fig:core-size} and \ref{fig:connectivity} report these quantities. Table~\ref{tab:prior} places the component counts in context.
\begin{table}[H]
\centering\small\setlength{\tabcolsep}{3pt}\renewcommand{\arraystretch}{1.3}
\begin{tabular}{@{}>{\raggedright\arraybackslash}p{2.7cm}>{\raggedright\arraybackslash}p{3.25cm}>{\raggedright\arraybackslash}p{3.4cm}>{\raggedright\arraybackslash}p{5.2cm}@{}}
\toprule
Study & Reported control scale & Body and validation & Relevance to the present study\\
\midrule
Collins et al. (2005) \cite{collins} & Passive dynamics with simple active control & Physical bipeds; level-ground walking & Body mechanics contribute directly to coordinated locomotion.\\
Liu et al. (2008) \cite{liucpg} & Six Matsuoka oscillators & Five-link biped; simulation & Compact rhythmic control with oscillator-level state dynamics.\\
Manoonpong et al. (2007) \cite{runbot} & Reflex network plus six additional adaptive neurons & Physical RunBot; boom-constrained walking and ramp adaptation & An adaptive neural layer works together with reflexes and body mechanics.\\
Y. Liu et al. (2024) \cite{eight} & Eight Stein-model neurons in the CPG & Go1 quadruped; physics simulation & Joint mappings, predefined foot paths, and fixed hip abduction/adduction define the control setting.\\
Steffen et al. (2026) \cite{steffen} & Reference MLP: $3\times128=384$ hidden units; spiking replacement uses ensembles & H1 humanoid; simulation; walking and arm-control selection & Distinguishes reference-network width from the neuronal ensembles used by the spiking implementation.\\
Radosavovic et al. (2023 version) \cite{radosavovic} & History-conditioned transformer & Full-size humanoid; physical outdoor locomotion & Provides a hardware-validated example of history-dependent humanoid control.\\
Present study & 3,609 recurrent core states; 101,263 nonzero recurrent weights & G1, 29 actuators / 15 policy outputs; physics simulation & Connectome-labelled core, explicit neural--body interfaces, and intervention-based pathway analysis.\\
\bottomrule
\end{tabular}
\caption{Selected primary-source context for control scale. Counts refer to the stated components. Task, embodiment, training, and unit dynamics differ across studies, so comparisons require component-level interpretation. The 2024 eight-neuron and 2026 spiking studies are cited in their arXiv versions.}
\label{tab:prior}
\end{table}
\FloatBarrier

\section{Complete condition summaries}\label{sec:conditions}
The recorded runtime is Node.js 24.19.0 and ONNX Runtime Web 1.30.0 on an Apple M2 running macOS 14.8.5, with the bundled MuJoCo WASM package reporting 3.10.0. The evaluator uses the package's observation and simulation modules. The protocol was archived before evaluation; a separate two-second smoke test checked execution. Wall-clock values include instrumentation and are execution diagnostics.

We compute planar body-frame velocity RMSE over non-terminated post-step observations:
\begin{equation}
E_v=\sqrt{\frac{1}{K}\sum_{t=1}^{K}\left[(v_{x,t}-v)^2+v_{y,t}^2\right]},
\end{equation}
Failed episodes have shorter windows, so these values are interpreted with termination time. All rows below aggregate three controlled initial-yaw offsets of a fixed checkpoint. Displacement and RMSE are medians for T\_graph, including failed conditions.
\small
\begin{longtable}{@{}lrrrrr@{}}
\caption{Complete terrain--speed summary. T\_graph and R1 entries are success counts over three yaw offsets.}\label{tab:conditions}\\
\toprule
Terrain & $v$ (m/s) & T\_graph & $\Delta x$ (m) & $E_v$ (m/s) & R1\\
\midrule\endfirsthead
\toprule Terrain & $v$ (m/s) & T\_graph & $\Delta x$ (m) & $E_v$ (m/s) & R1\\\midrule\endhead
\ConditionRows
\bottomrule
\end{longtable}
\normalsize

\section{State-matched visual evidence}
Eleven replays visualize existing nominal-yaw episodes: all seven intact 0.5 m/s terrains, the flat-ground 0.5 m/s reset, rough and discrete steps at 0.8 m/s, and ascending stairs at 0.3 m/s. All endpoints, termination flags, and step counts exactly match the original records on the same runtime. The 115 archived states include joint positions, geometry transforms, and core activations. The three.js viewer renders those poses; camera, lighting, colors, arrows, and labels are presentation settings. All 28 distinct screenshots are retained across the main text and supplement.

\begin{figure}[htbp]
\centering
\shot{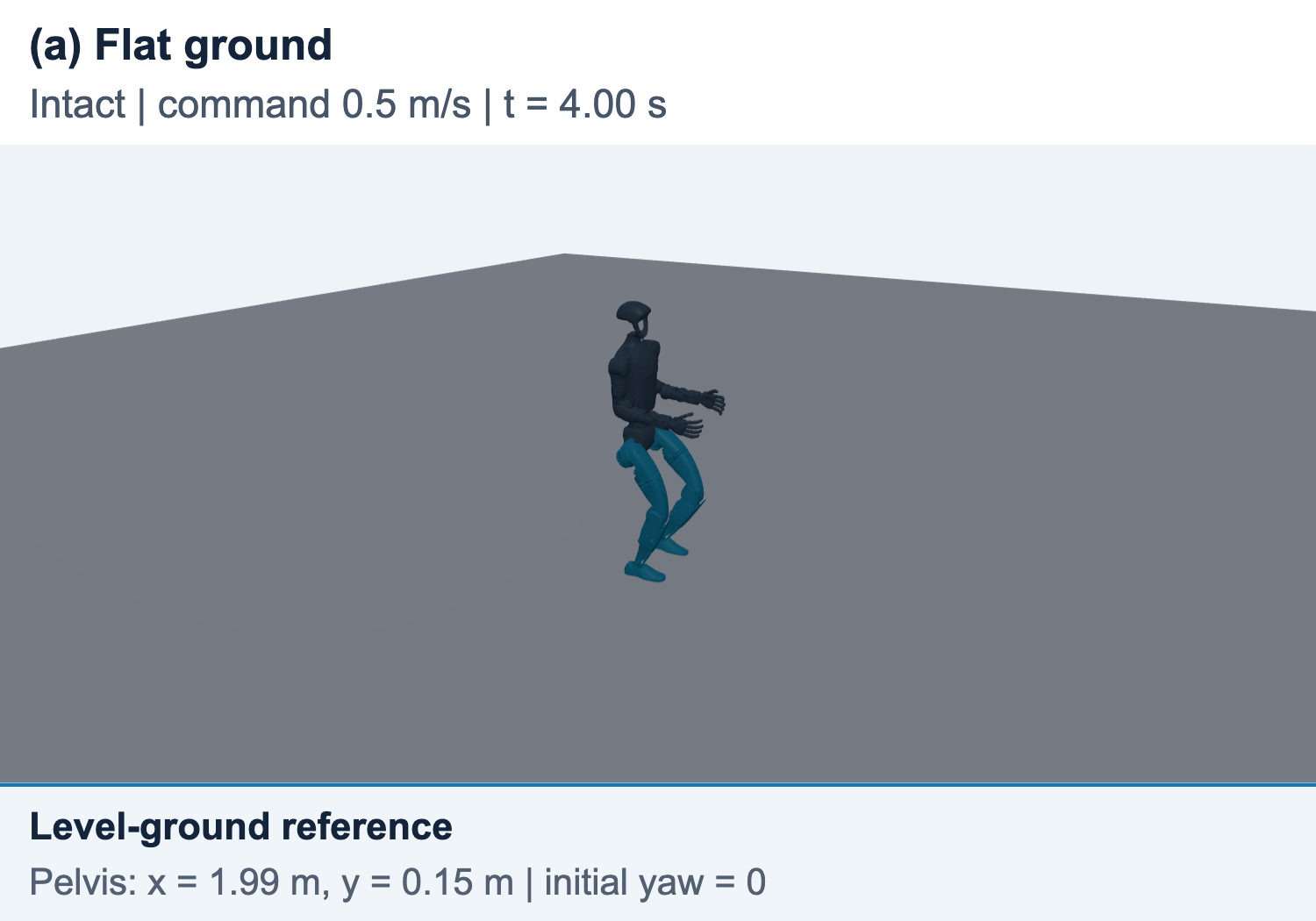}\hfill\shot{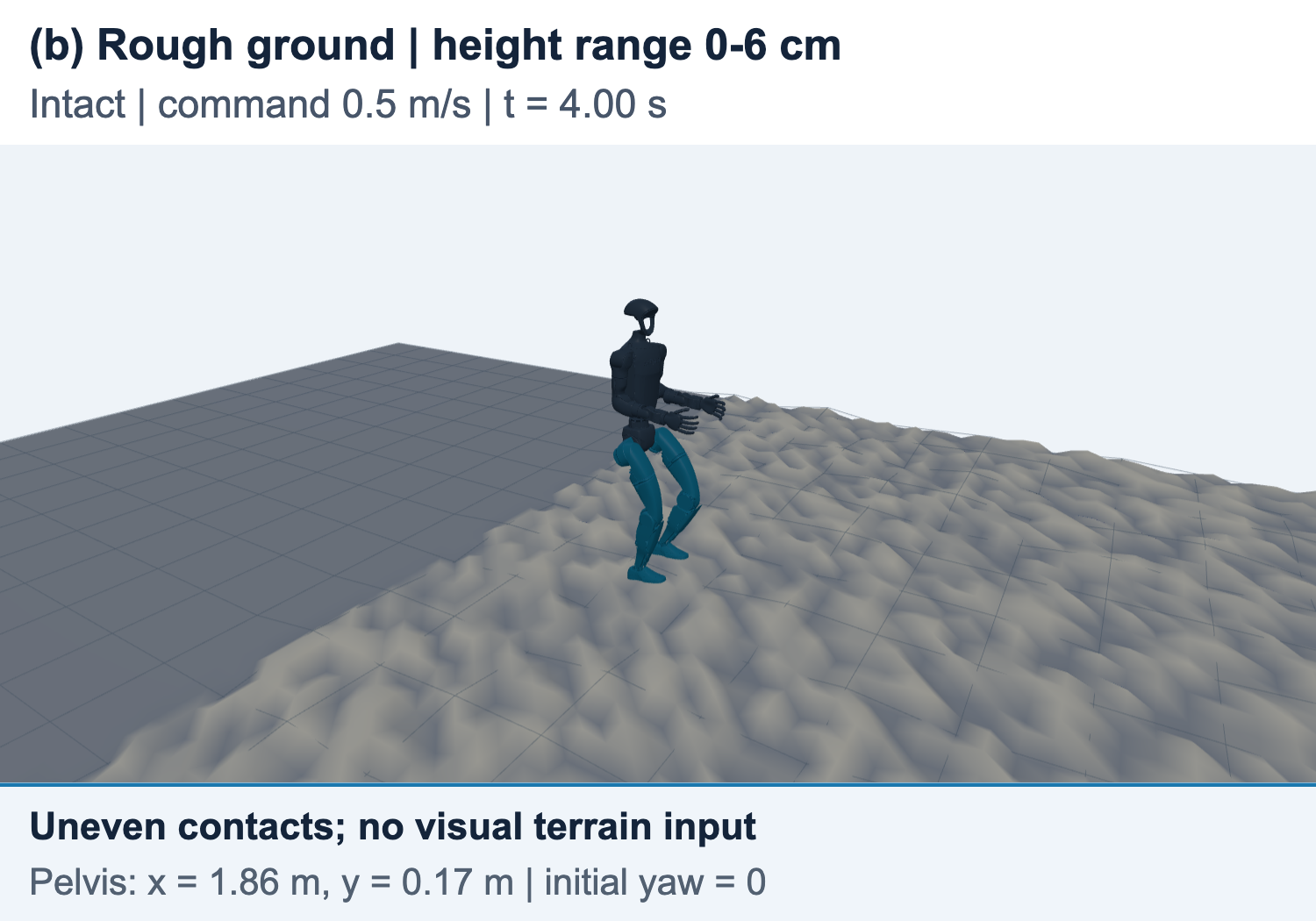}\par\vspace{3pt}
\shot{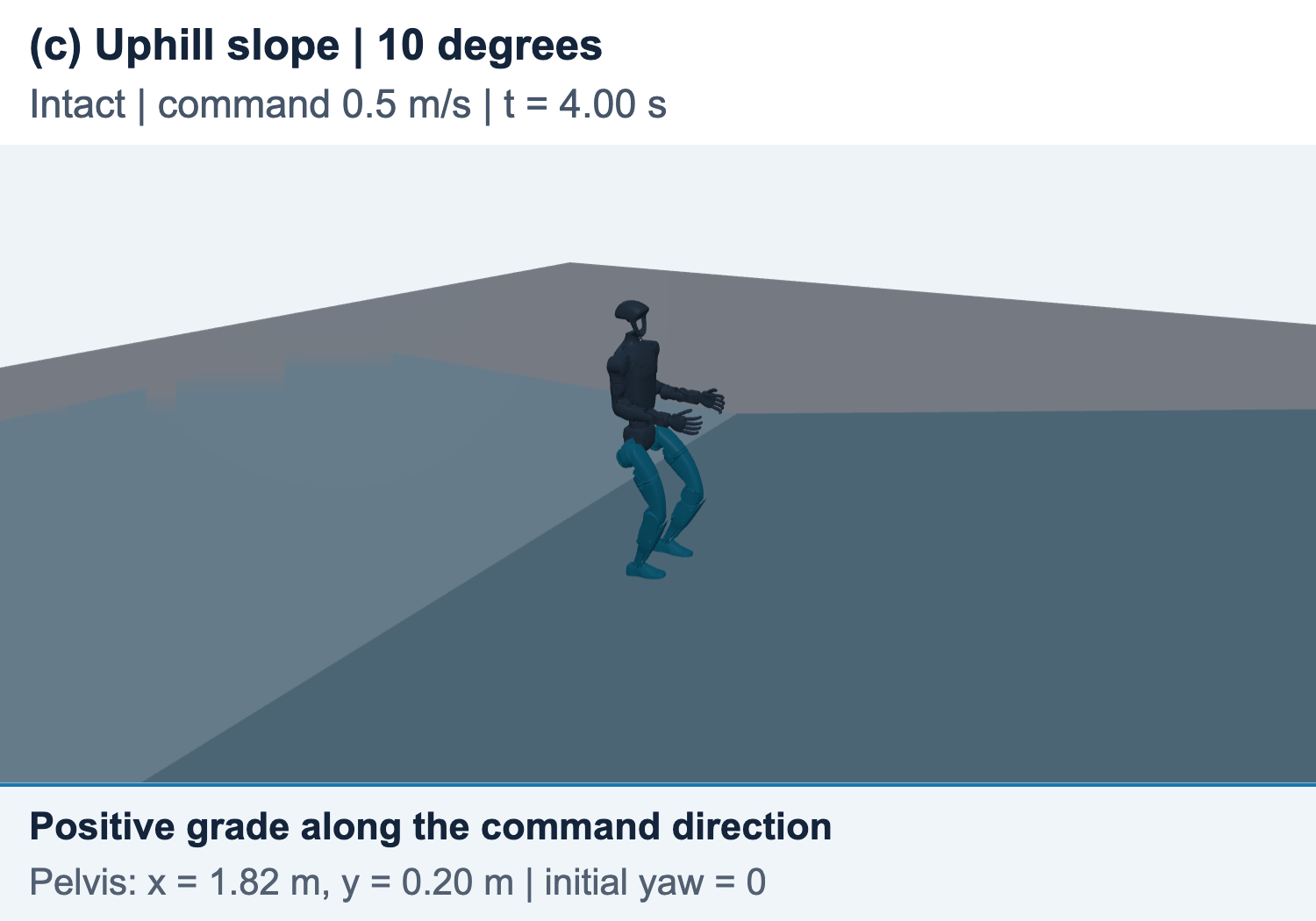}\hfill\shot{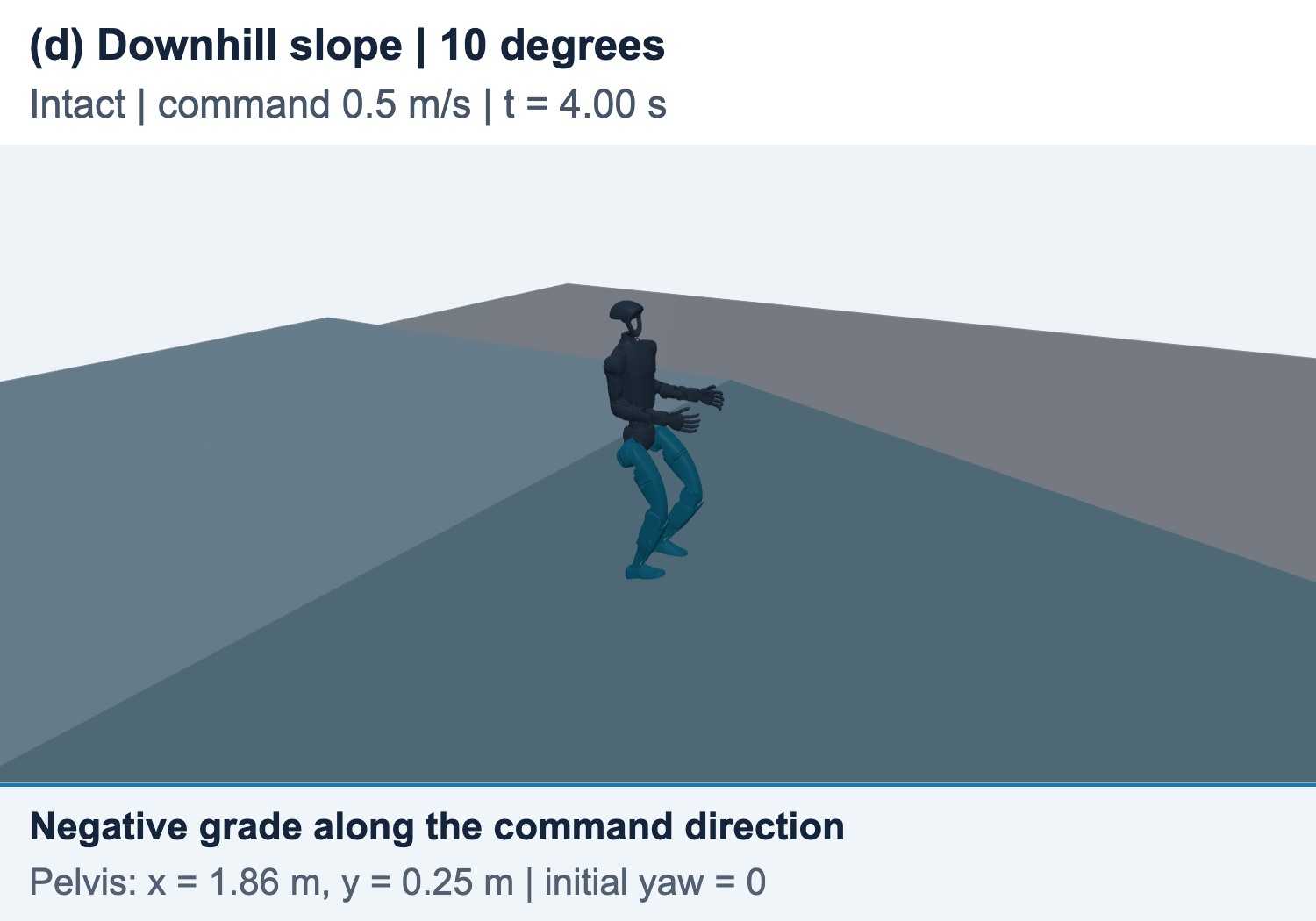}\par\vspace{3pt}
\shot{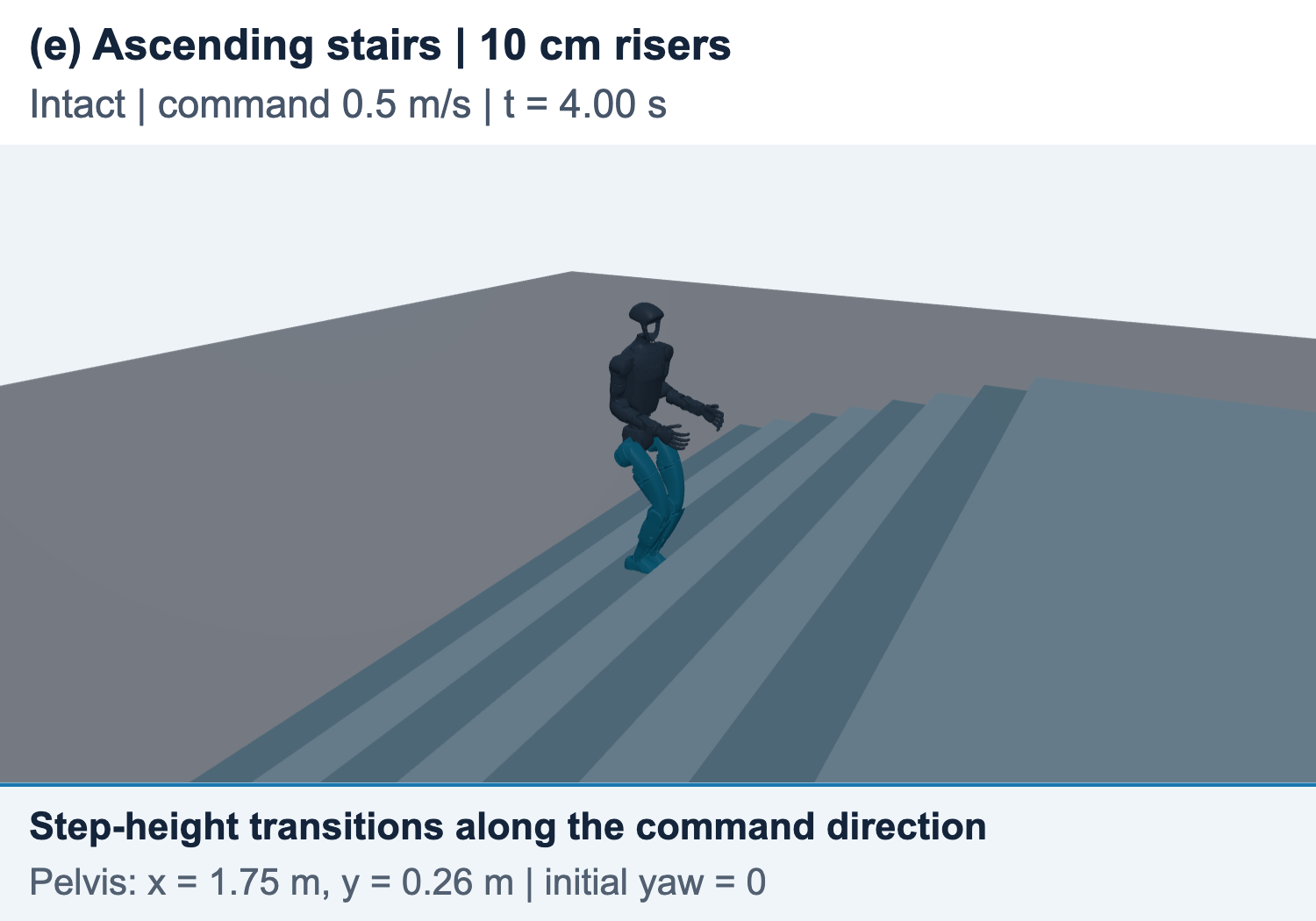}\hfill\shot{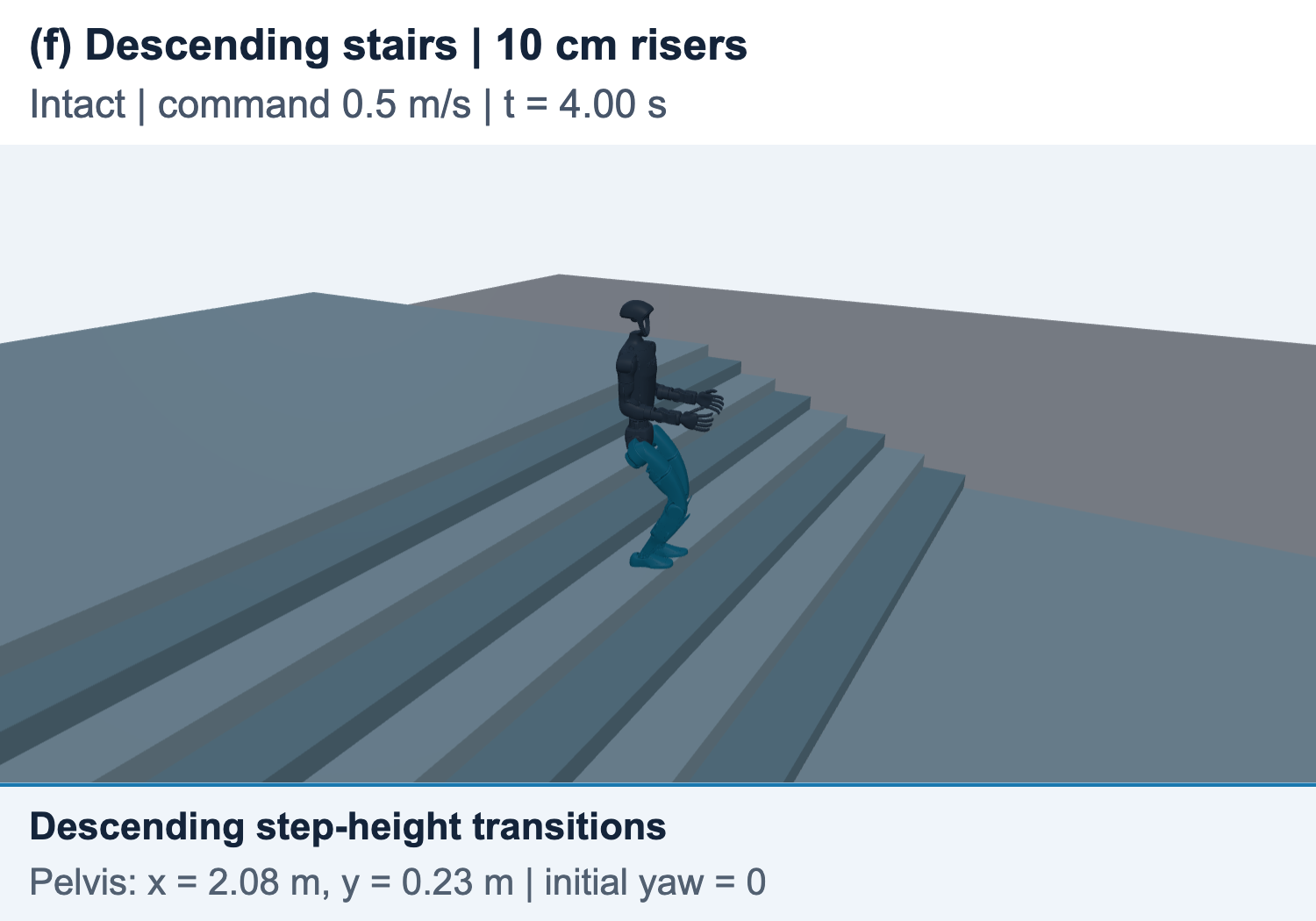}\par\vspace{3pt}
\shot{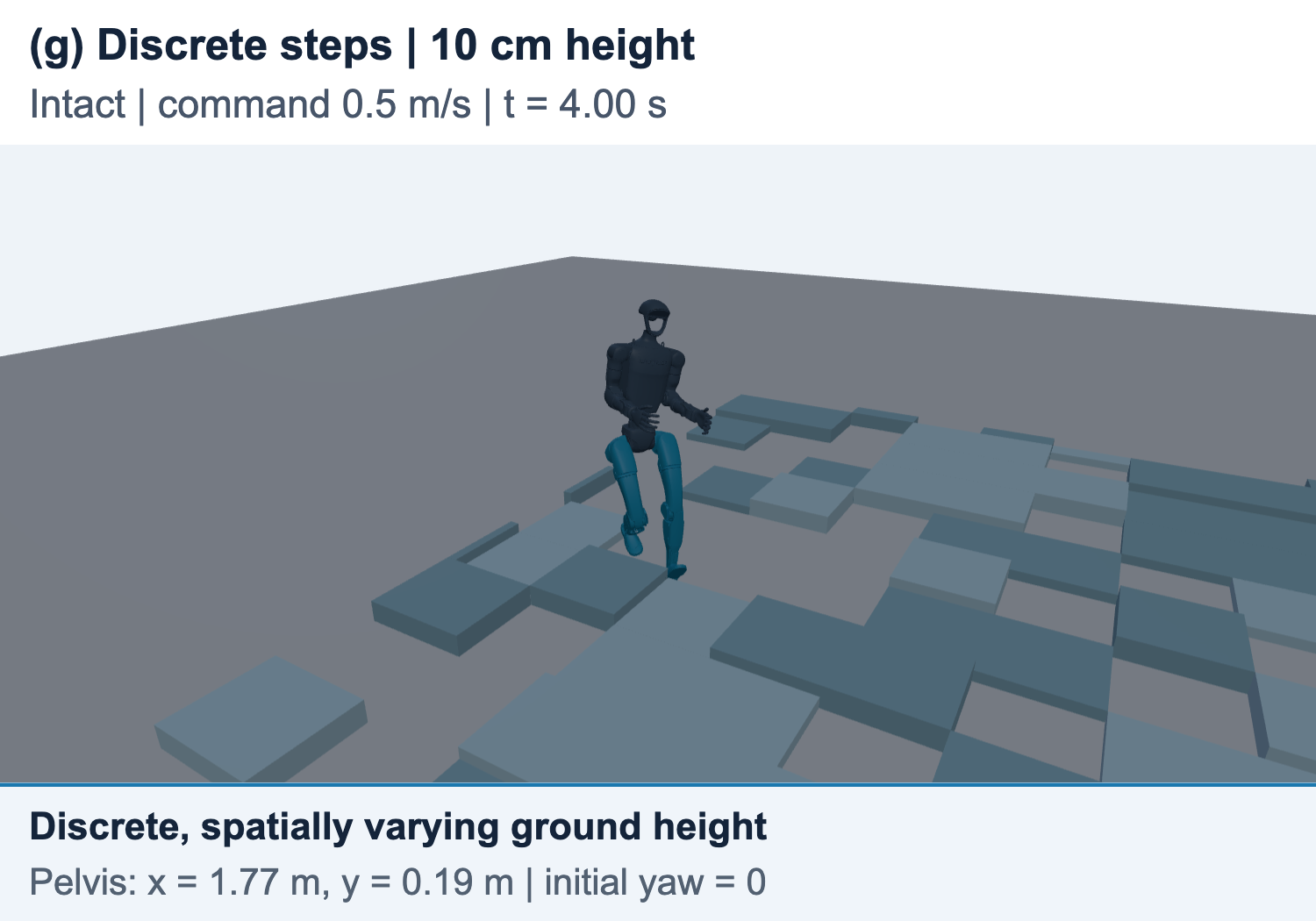}\hfill\includegraphics[width=.485\linewidth]{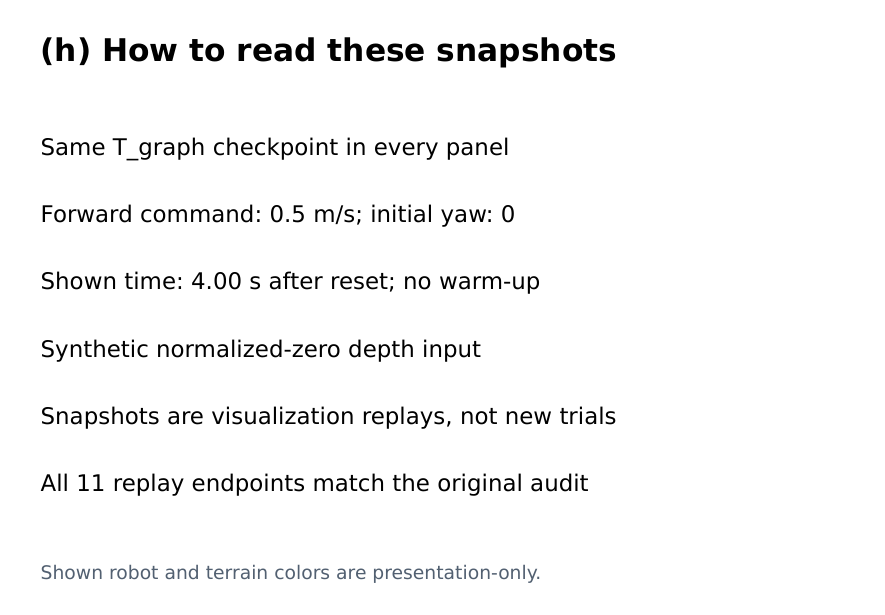}
\caption{Annotated terrain atlas from exact-state simulation replays. (a--g) The same intact checkpoint, 0.5 m/s command, zero initial yaw, and 4.00 s elapsed time on flat, rough, uphill, downhill, upstairs, downstairs, and discrete-step terrain. (h) Shared reading protocol. External-view renders show the terrain geometry and recorded robot pose in each test environment.}
\label{fig:atlas}
\end{figure}
\begin{figure}[htbp]
\centering\shot{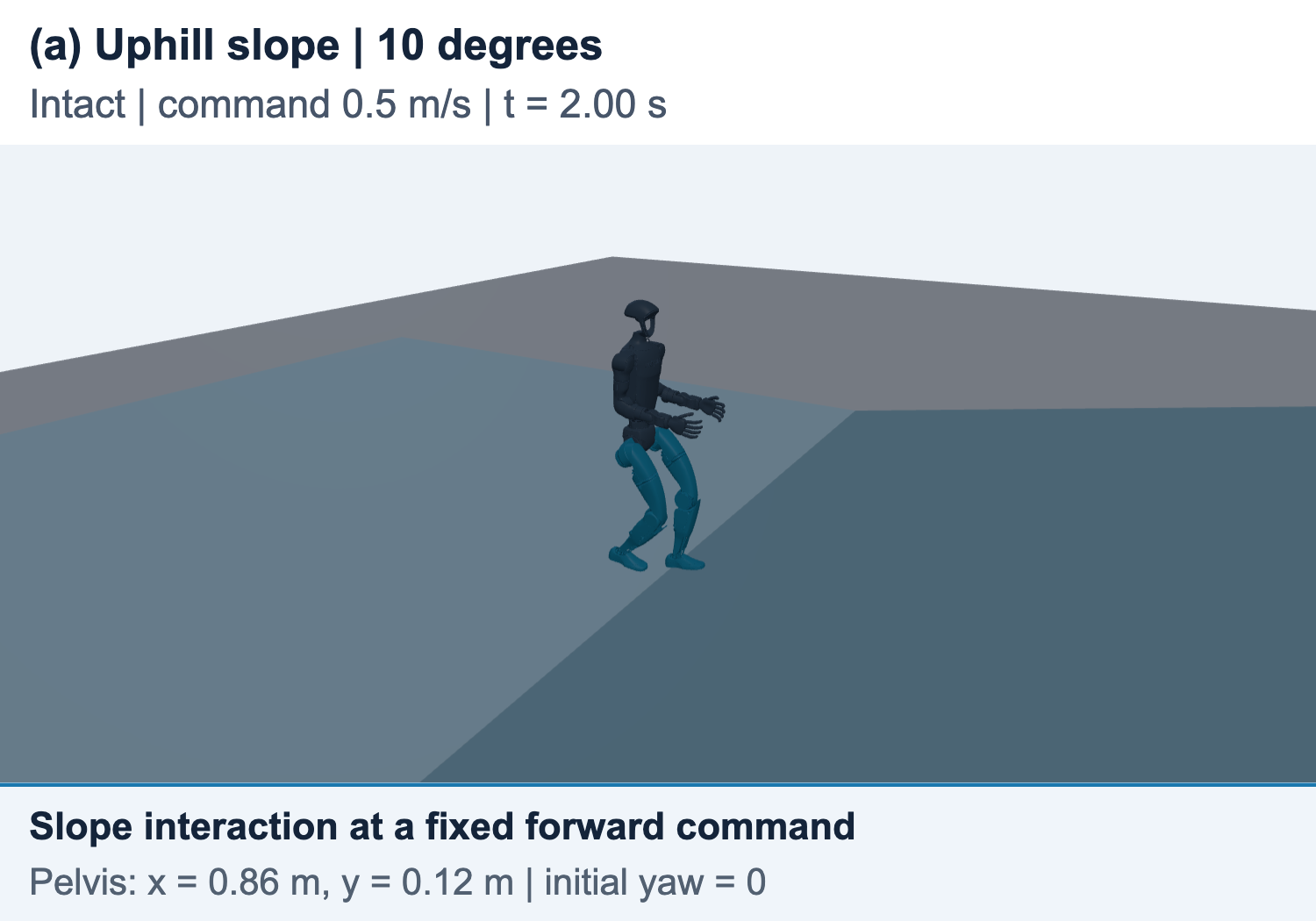}\hfill\shot{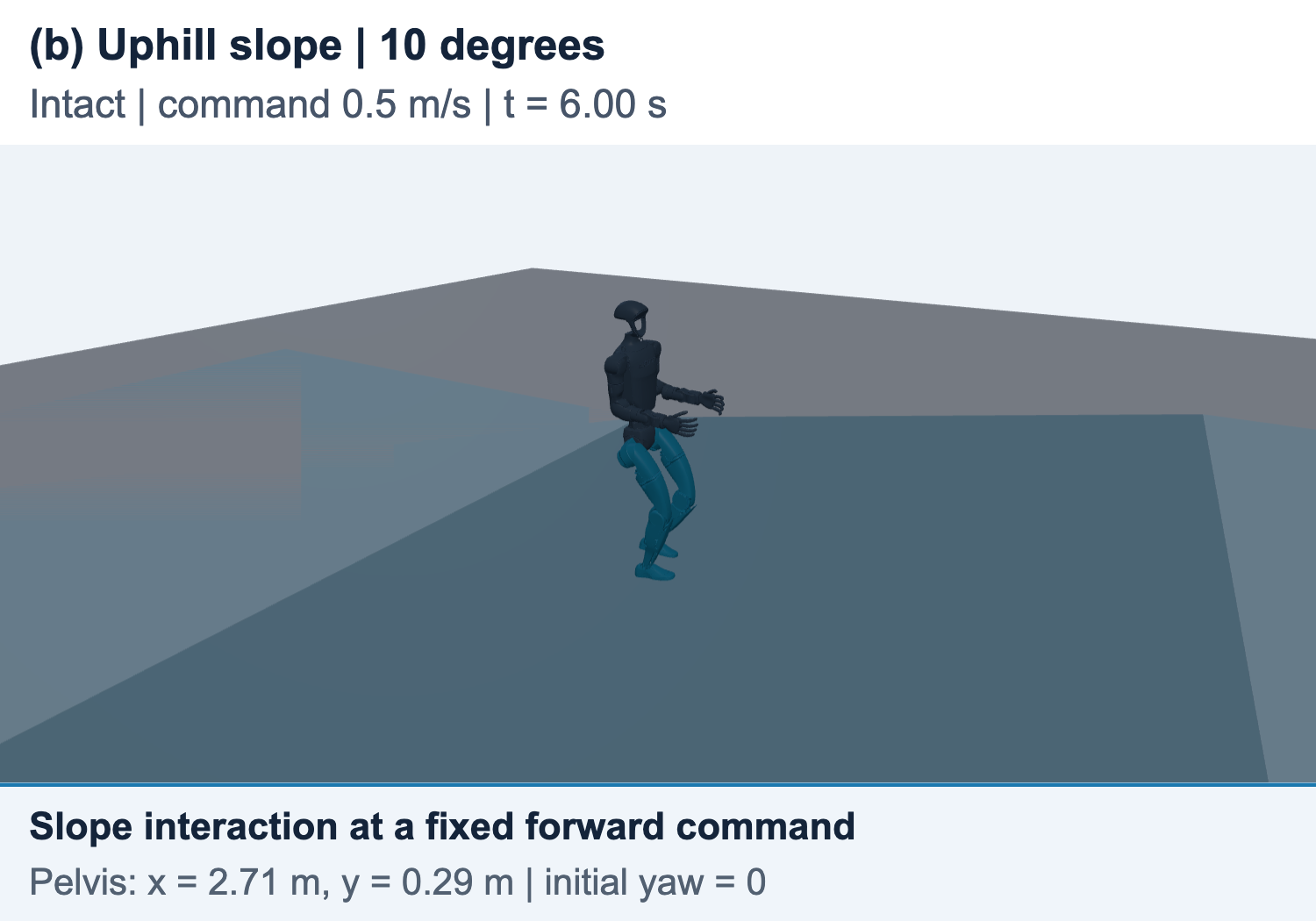}\par\vspace{4pt}
\shot{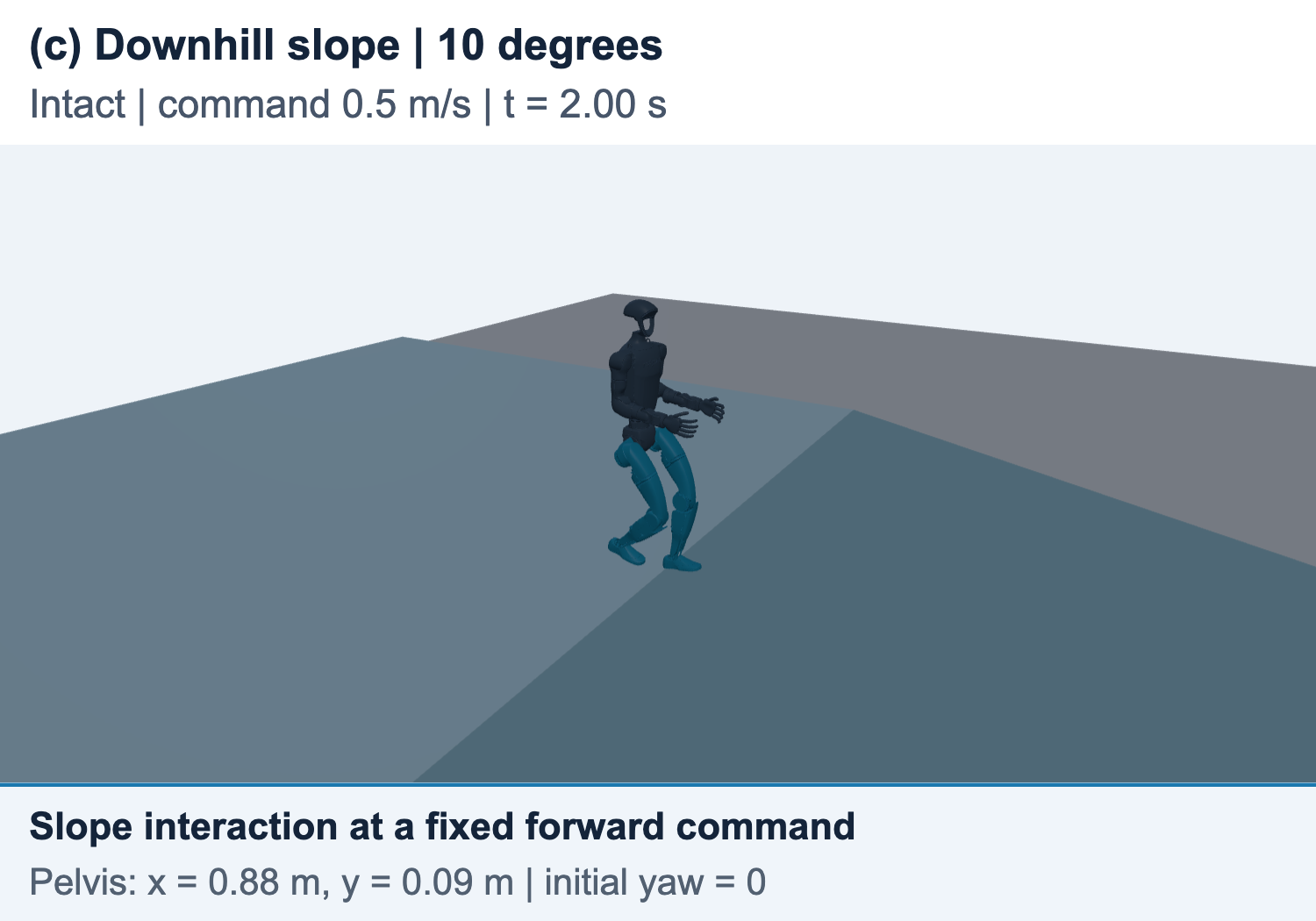}\hfill\shot{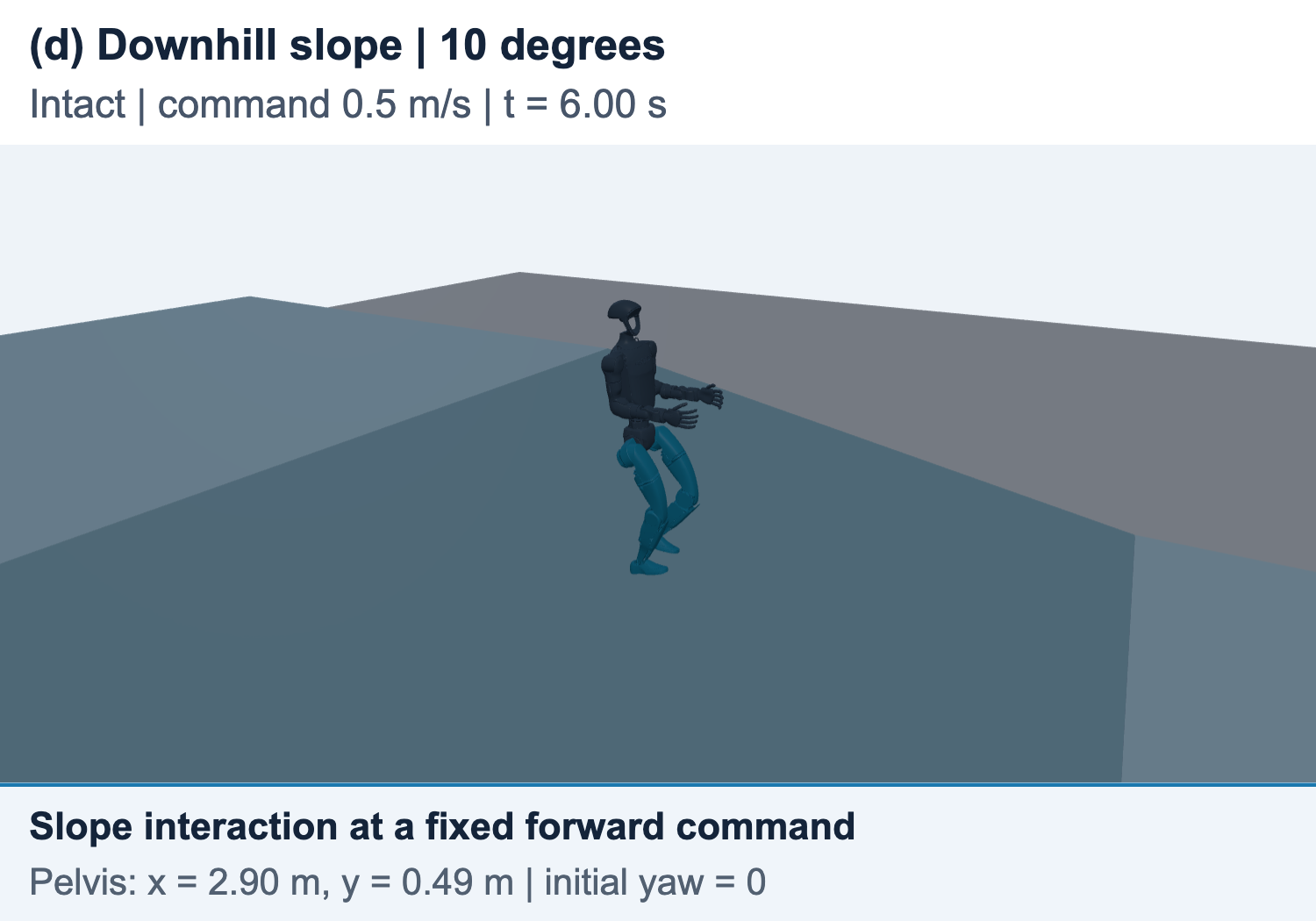}
\caption{Slope interaction at a fixed 0.5 m/s command. Top: uphill $10^\circ$ at 2 and 6 s; bottom: downhill $10^\circ$ at the same elapsed times. Geometry, pose, and reported coordinates are recorded simulation quantities. The two separate episodes show distinct posture and displacement under a common command, with different contact and gravity conditions.}
\label{fig:slope-seq}
\end{figure}
\begin{figure}[htbp]
\centering\shot{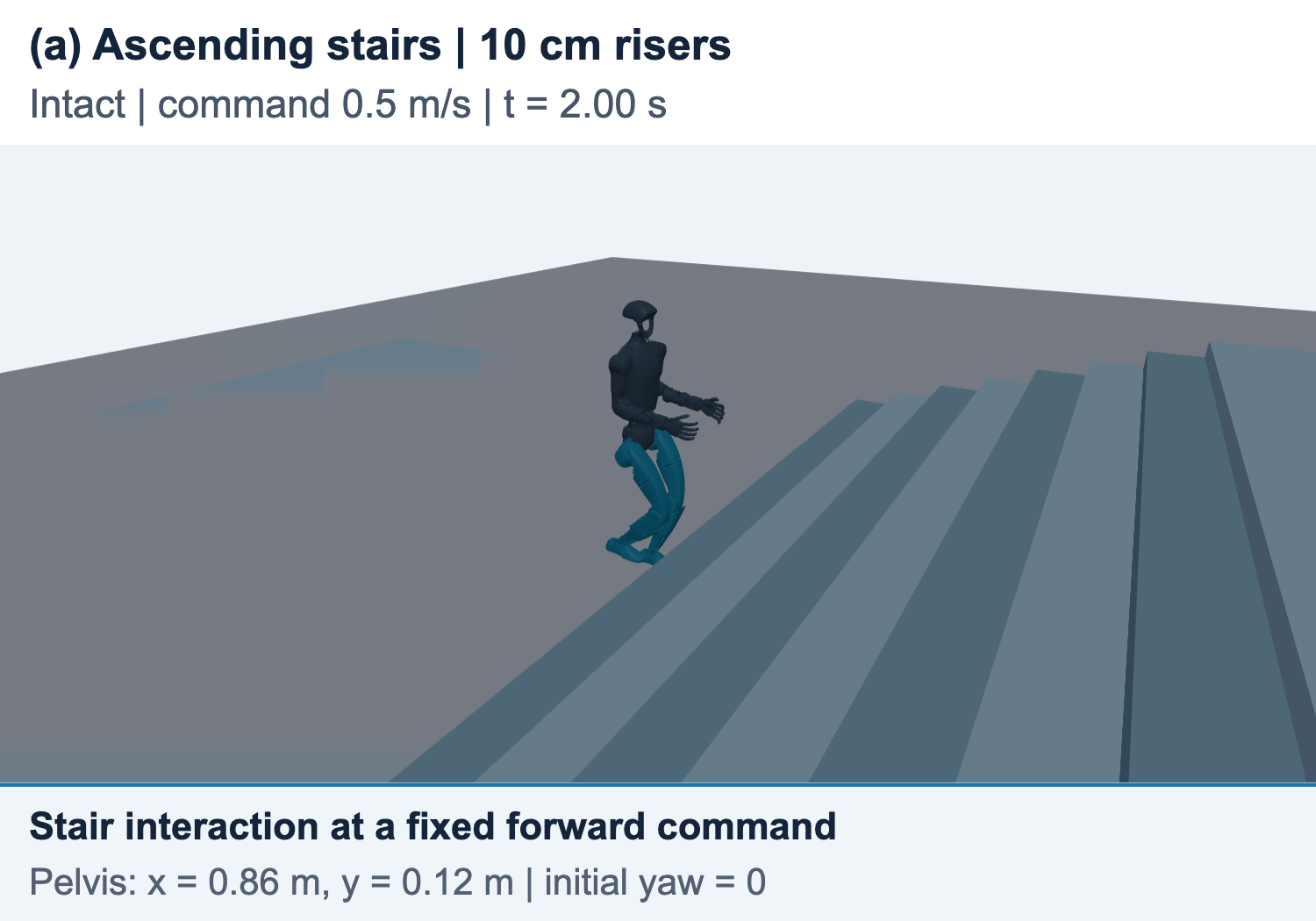}\hfill\shot{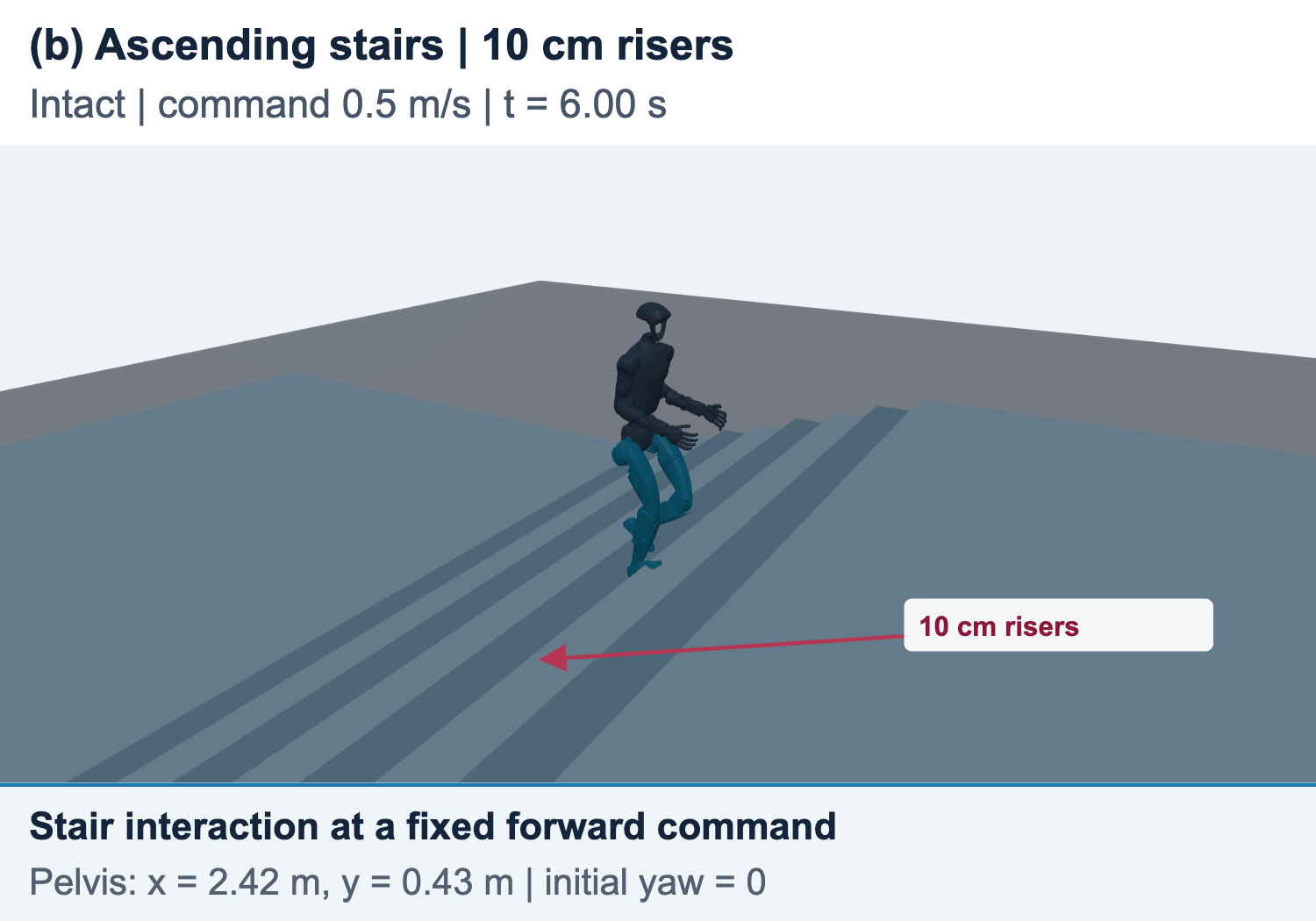}\par\vspace{4pt}
\shot{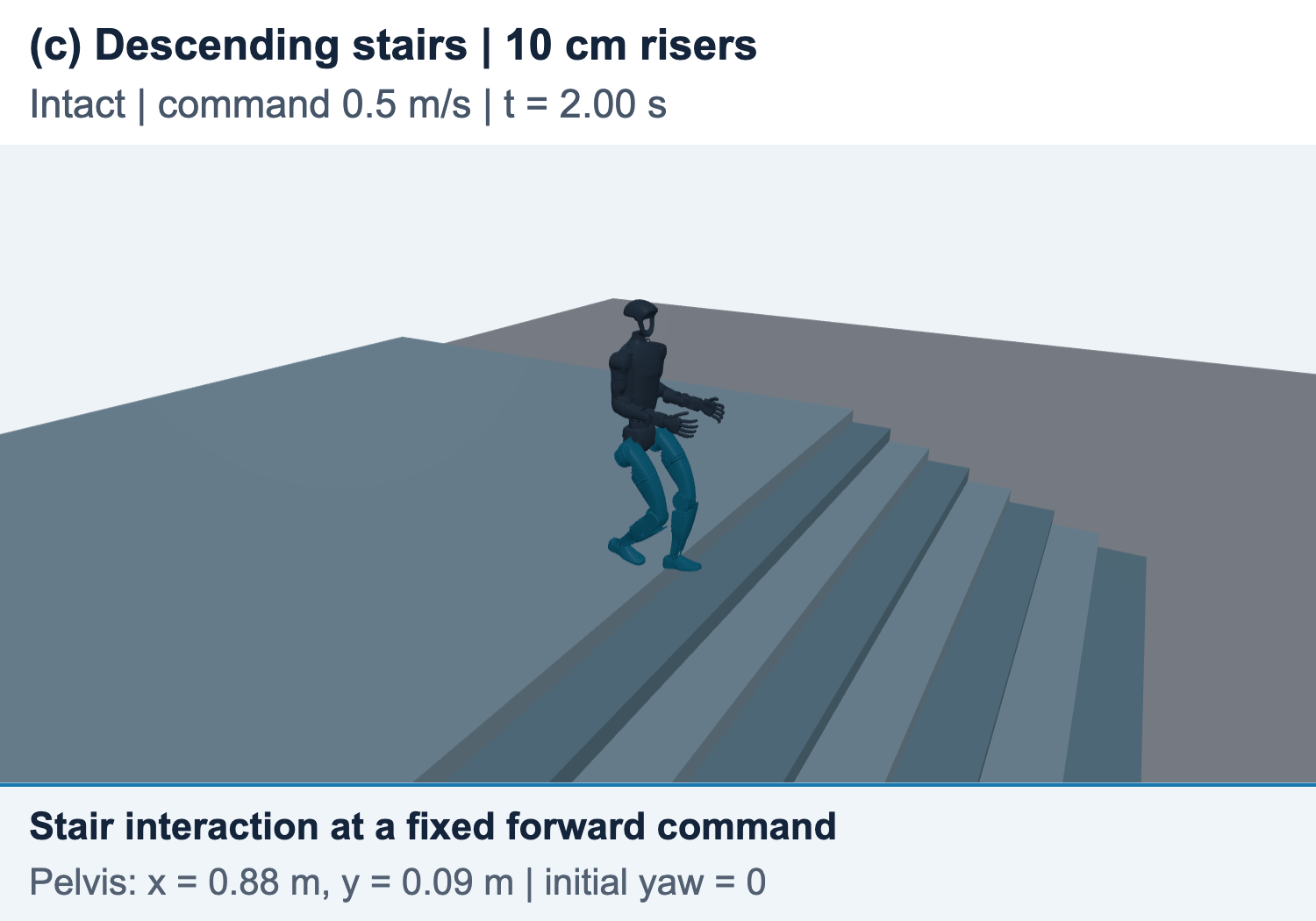}\hfill\shot{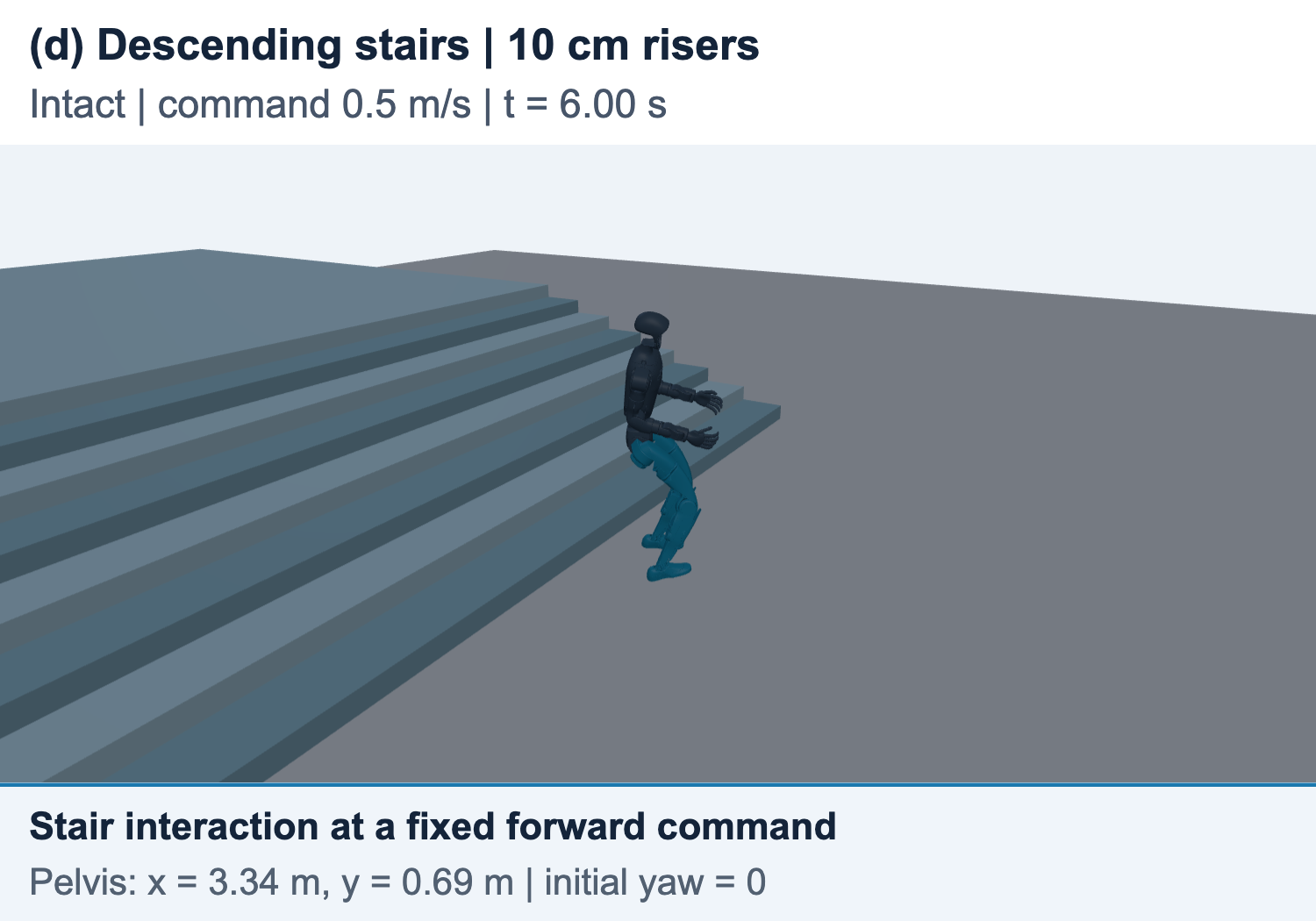}
\caption{Stair interaction at a fixed 0.5 m/s command. Top: ascending 10 cm stairs at 2 and 6 s; bottom: descending 10 cm stairs at the same elapsed times. The annotation marks riser geometry. Recorded pose and coordinates show the movement attained in these two episodes under identical command settings.}
\label{fig:stair-seq}
\end{figure}
\begin{figure}[htbp]
\centering
\includegraphics[width=.575\linewidth]{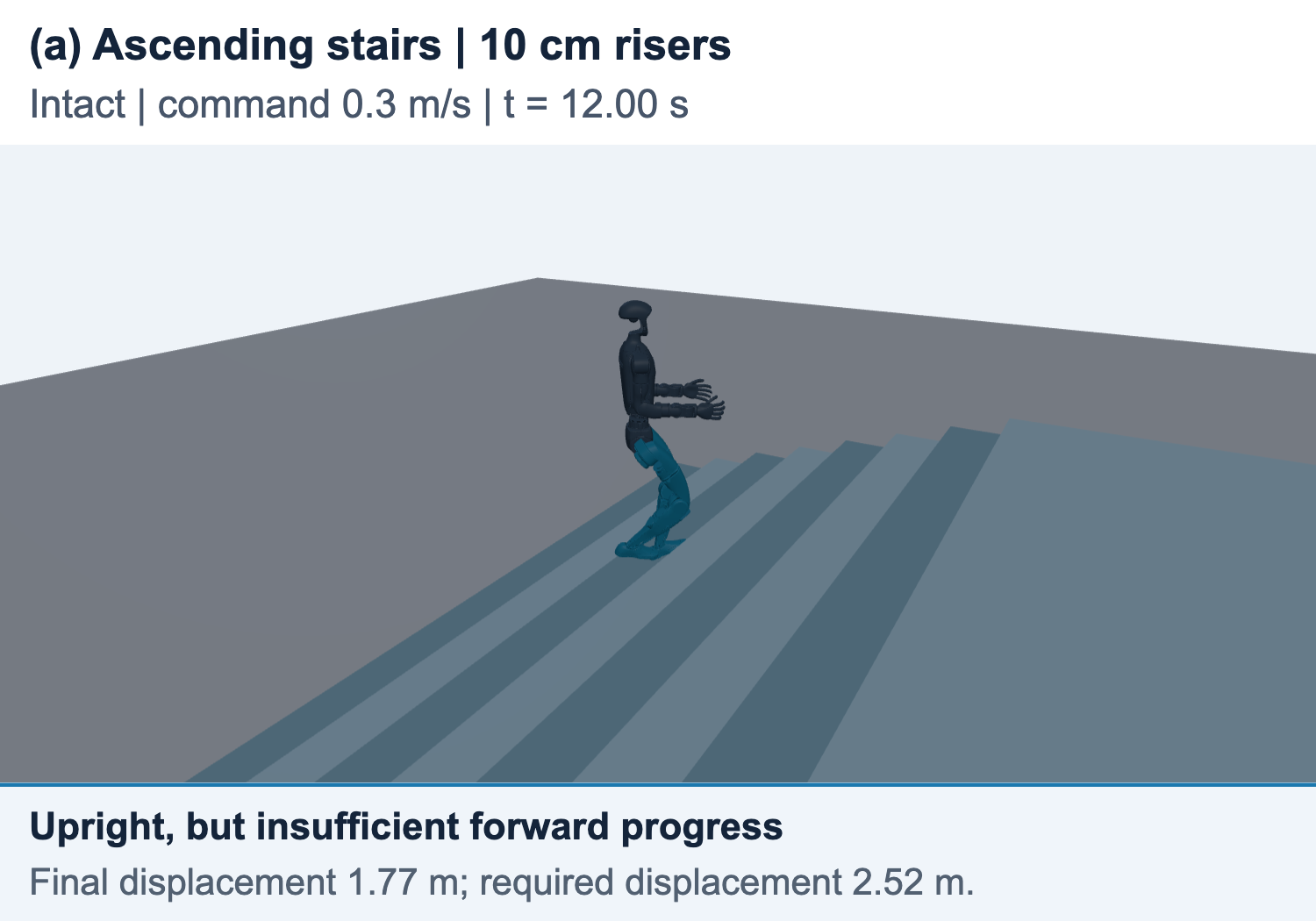}\hfill\includegraphics[width=.405\linewidth]{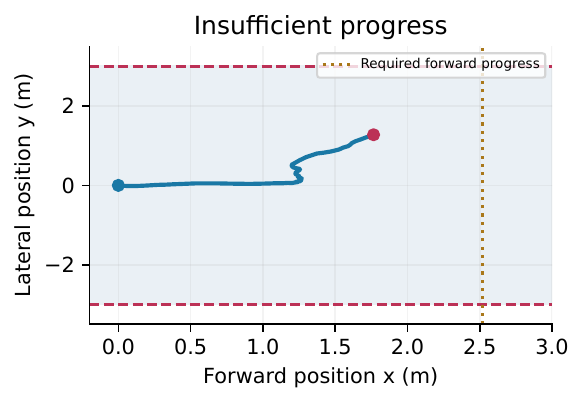}\par\vspace{4pt}
\includegraphics[width=.575\linewidth]{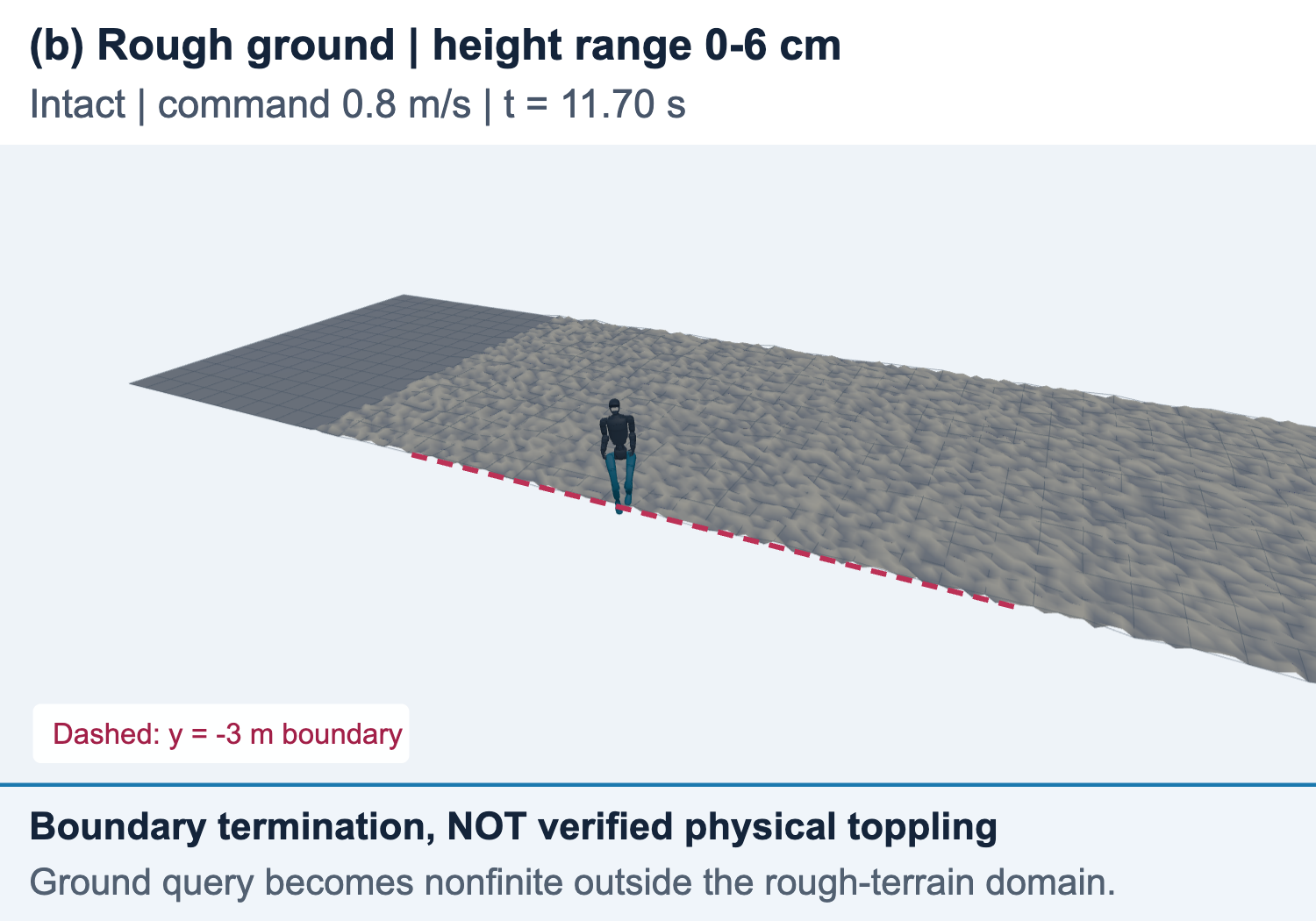}\hfill\includegraphics[width=.405\linewidth]{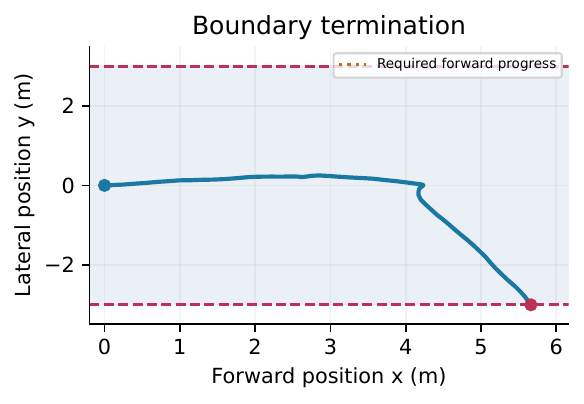}\par\vspace{4pt}
\includegraphics[width=.575\linewidth]{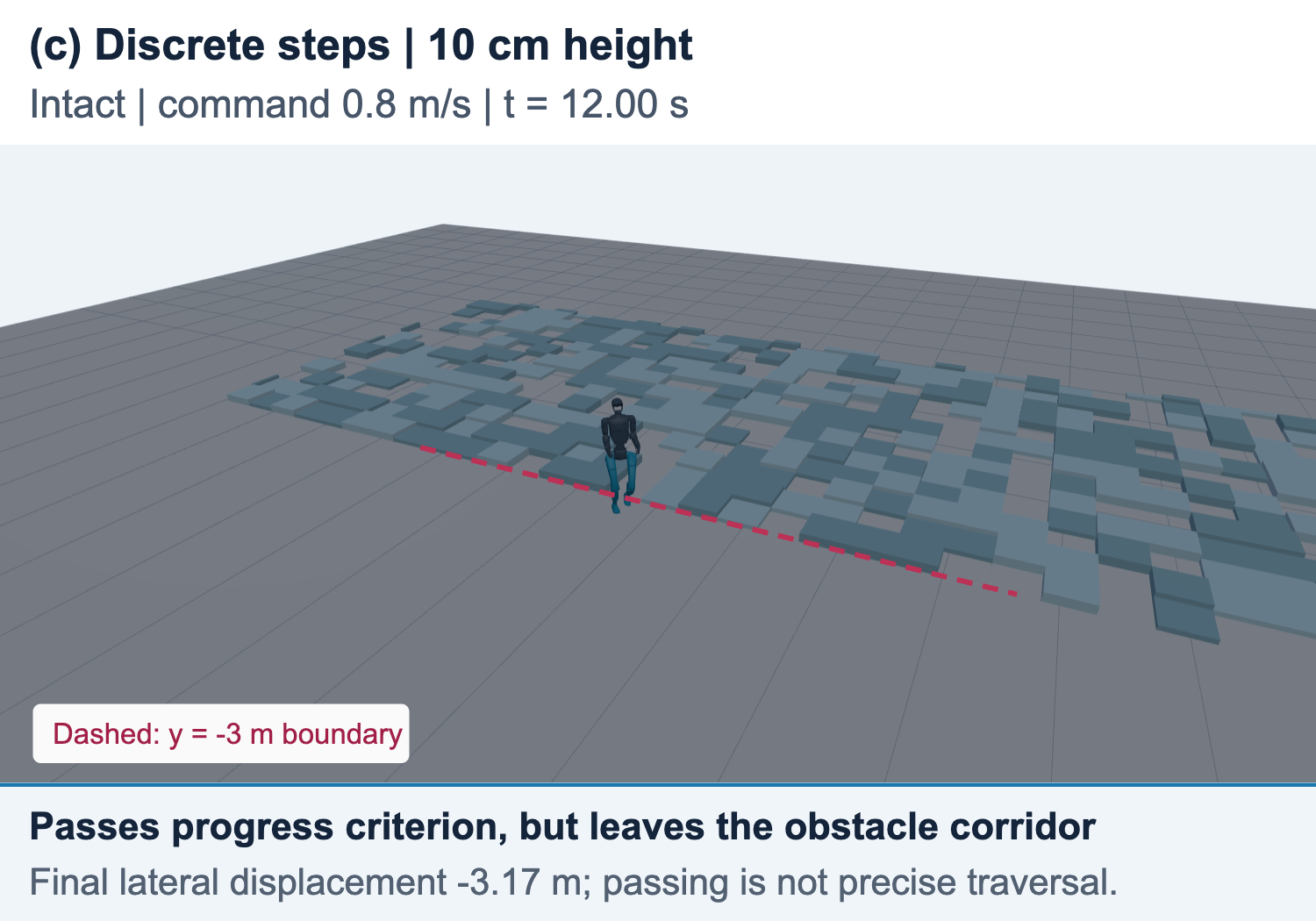}\hfill\includegraphics[width=.405\linewidth]{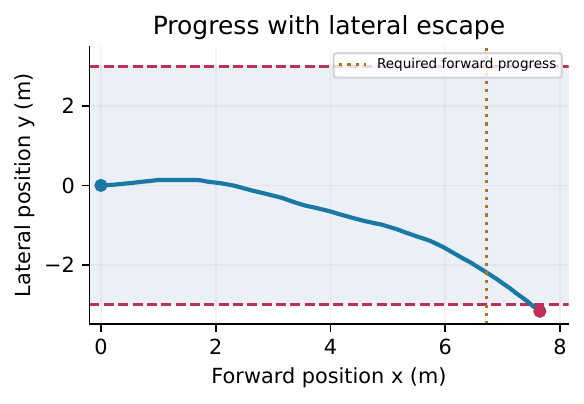}
\caption{Termination and trajectory-deviation cases, all at zero initial yaw. (a) Upstairs, 0.3 m/s: upright at 12 s but below the progress threshold. (b) Rough, 0.8 m/s: the episode terminates at 11.70 s after leaving the valid ground-height domain. (c) Discrete steps, 0.8 m/s: the 12 s progress criterion passes despite lateral escape. Right: sampled trajectories with exact final endpoints; dashed lines mark $y=\pm3$ m, dotted lines mark required forward displacement, and red dots mark endpoints. Screenshot dashed lines project the actual $y=-3$ m boundary.}
\label{fig:failure-images}
\end{figure}
\begin{figure}[htbp]
\centering\includegraphics[width=\linewidth]{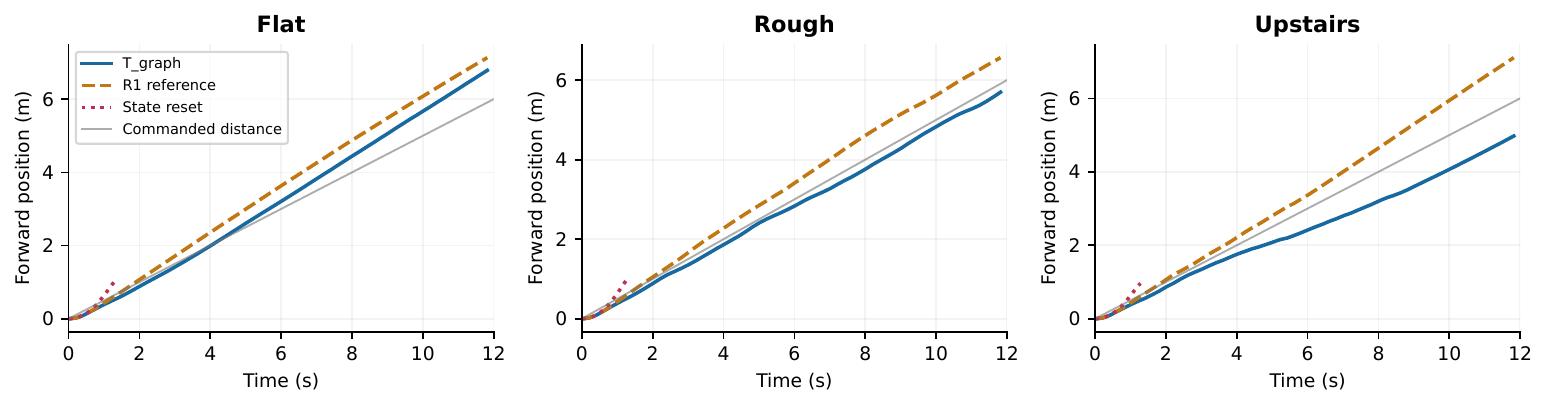}
\caption{Forward position for the fixed 0.5 m/s, nominal-yaw conditions on flat, rough, and ascending-stair terrains. Colored curves show recorded position; the gray line gives the commanded distance $vt$. Traces end at termination. These examples complement the complete condition-level outcome table.}
\label{fig:traces}
\end{figure}
\FloatBarrier

\section{Recorded core activity}\label{sec:activity}
Group means summarize artificial activations in the intact 0.5 m/s rough-ground episode. The DN-labelled core states participate in the engineered recurrence and can be active while the separate descending bottleneck is zero.
\begin{figure}[htbp]
\centering\includegraphics[width=\linewidth]{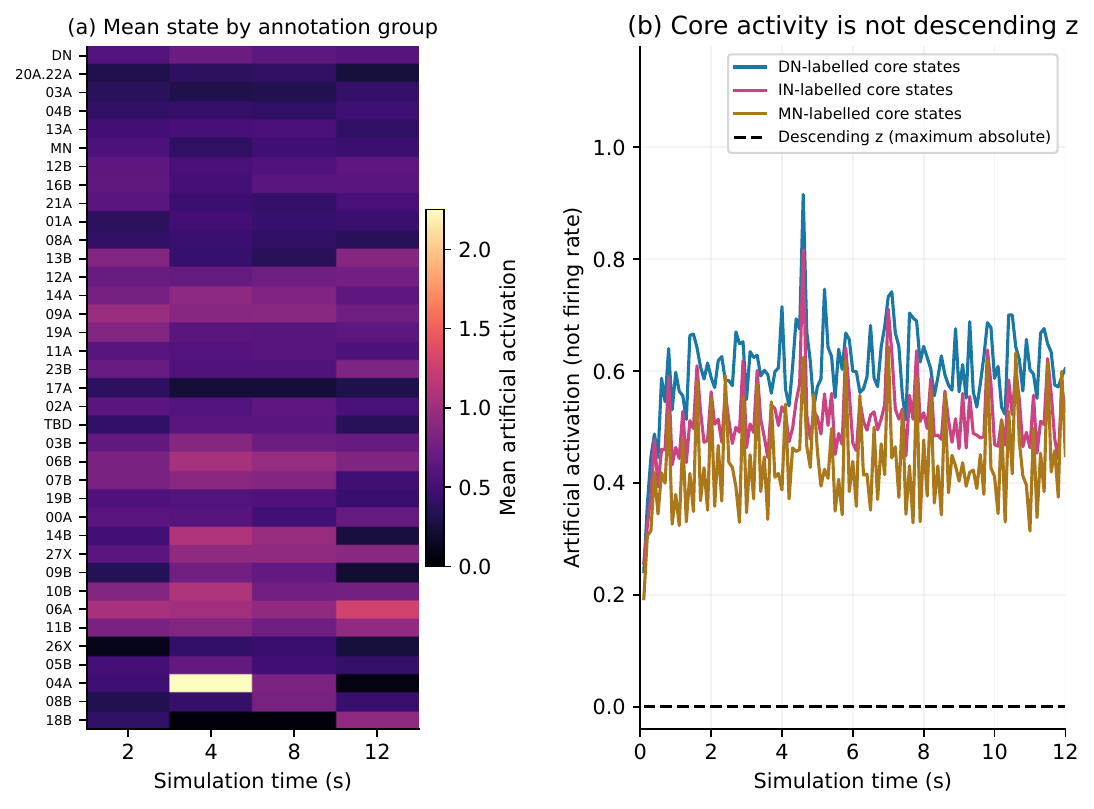}
\caption{Recorded artificial core activity during the same 0.5 m/s rough-ground episode as Figure~\ref{fig:rough-seq}. (a) Mean activation within each supplied anatomical annotation group at the four screenshot times, on one shared linear color scale. (b) Class-mean core activations sampled every 0.1 s, contrasted with the zero descending bottleneck. The DN-labelled entries belong to the recurrent core and receive its engineered drive, whereas $z$ is the separate upstream branch output.}
\label{fig:neural}
\end{figure}
\FloatBarrier

\section{Provenance and reproducibility records}\label{sec:provenance}
The inspected package contains the T\_graph ONNX checkpoint, deployment contract, robot model, terrain descriptions, and \texttt{t1\_w5} neuron annotations. Its metadata identifies the network as a distilled student and attributes the annotations to MaleCNS v1.0. The original graph-extraction code, selected biological adjacency, teacher lineage, training objectives and schedules, and training/validation splits were unavailable in the supplied materials. The filtering meaning of \texttt{w5} is undocumented. Accordingly, anatomical provenance is attributed to package metadata, while computation and model-level counts are reconstructed from the deployed export.

Completing the record requires the source dataset snapshot, rules for cell selection and synaptic filtering, graph normalization, declarations of fixed and trainable parameters, and the training code and configuration. Public redistribution of the separate inference bundle also requires confirmed terms for its code, robot assets, and weights. The author-held archive contains evaluation records and analysis scripts; a full evaluation rerun additionally requires that bundle. Legacy success-rate metadata without trial records is excluded.

As a numerical implementation check, the archived six diagnostic calls compare float64 algebra with float32 ONNX outputs: maximum core-state error is $1.69\times10^{-6}$ and raw-action error is $1.22\times10^{-6}$, with exact recorded joint-target mappings. The observations, servo parameters, and model hashes are retained in the archive. These short checks validate the implemented equations at those inputs and are separate from the 147 study episodes.